\documentclass[sigconf,screen]{acmart}

\setcopyright{rightsretained}
\acmPrice{}
\acmDOI{10.1145/3575693.3575702}
\acmYear{2023}
\copyrightyear{2023}
\acmSubmissionID{asplosb23main-p47-p}
\acmISBN{978-1-4503-9916-6/23/03}
\acmConference[ASPLOS '23]{Proceedings of the 28th ACM International Conference on Architectural Support for Programming Languages and Operating Systems, Volume 2}{March 25--29, 2023}{Vancouver, BC, Canada}
\acmBooktitle{Proceedings of the 28th ACM International Conference on Architectural Support for Programming Languages and Operating Systems, Volume 2 (ASPLOS '23), March 25--29, 2023, Vancouver, BC, Canada}
\received{2022-07-07}
\received[accepted]{2022-09-22}



\newcommand{\asplossubmissionnumber}{47}

\newcommand*\circled[1]{\tikz[baseline=(char.base)]{\node[shape=circle,fill,inner sep=0.5pt] (char) {\small{\textcolor{white}{#1}}};}}

\usepackage[normalem]{ulem}
\usepackage{amsfonts}
\usepackage{amsmath}
\usepackage{xspace}
\usepackage{boldline} %
\usepackage{tikz}  %
\usepackage[utf8]{inputenc}
\usepackage[T1]{fontenc}
\usepackage{microtype}
\usepackage{multicol}

\begin{document}

\newcommand{\sysName}[0]{Hidet}

\newcommand{\code}[1]{\texttt{#1}}
\newcommand{\system}[0]{\sysName\xspace}
\hyphenation{\sysName}
\newcommand{\AvgSpeedup}[0]{$1.22\times$\xspace} 
\newcommand{\MaxSpeedup}[0]{$1.48\times$\xspace}
\newcommand{\TuningTimeReduction}[0]{$11\times$\xspace}
\newcommand{\AutoTVMTuningTimeReduction}[0]{$20\times$\xspace}
\newcommand{\DNNCompilers}[0]{tiramisu,tvm,halide,tensorflow-xla,flextensor}
\newcommand{\revise}[1]{#1}

\settopmatter{printfolios=true}

\title{Hidet:  Task-Mapping Programming Paradigm for Deep Learning Tensor Programs}

\author{Yaoyao Ding}
\authornote{Part of the work done while interning at Amazon.}
\authornote{Also with Vector Institute.}
\affiliation{%
  \institution{University of Toronto}
  \city{Toronto}
  \country{Canada}
}
\email{yaoyao@cs.toronto.edu}

\author{Cody Hao Yu}
\affiliation{%
  \institution{Amazon Web Services}
  \city{Santa Clara}
  \country{USA}
}
\email{hyuz@amazon.com}

\author{Bojian Zheng}
\authornotemark[2]
\affiliation{%
  \institution{University of Toronto}
  \city{Toronto}
  \country{Canada}
}

\email{bojian@cs.toronto.edu}

\author{Yizhi Liu}
\affiliation{%
  \institution{Amazon Web Services}
  \city{Santa Clara}
  \country{USA}
}
\email{yizhiliu@amazon.com}

\author{Yida Wang}
\affiliation{%
  \institution{Amazon Web Services}
  \city{Santa Clara}
  \country{USA}
}
\email{wangyida@amazon.com}

\author{Gennady Pekhimenko}
\authornotemark[2]
\affiliation{%
  \institution{University of Toronto}
  \city{Toronto}
  \country{Canada}
}
\email{pekhimenko@cs.toronto.edu}

\begin{abstract}

As deep learning models nowadays are widely adopted by both cloud services and edge devices, reducing the latency of deep learning model inferences becomes crucial to provide efficient model serving.
However, it is challenging to develop efficient tensor programs for deep learning operators due to the high complexity of modern accelerators (e.g., NVIDIA GPUs and Google TPUs) and the rapidly growing number of operators.

Deep learning compilers, such as Apache TVM, adopt declarative scheduling primitives to lower the bar of developing tensor programs. 
However, we show that this approach is insufficient to cover state-of-the-art tensor program optimizations (e.g., double buffering).
In this paper, we propose to embed the scheduling process into tensor programs and use dedicated mappings, called task mappings, to define the computation assignment and ordering directly in the tensor programs. 
This new approach greatly enriches the expressible optimizations by allowing developers to manipulate tensor programs at a much finer granularity (e.g., allowing program-statement-level optimizations).
We call the proposed method the task-mapping programming paradigm. 
In addition, we propose a new post-scheduling fusion optimization that allows developers to focus on scheduling every single operator and automates the fusion after scheduling. 
It greatly reduces the engineering efforts for operator fusion.
Our proposed paradigm also constructs an efficient hardware-centric schedule space, which is agnostic to the program input size and greatly reduces the tuning time. 

With the proposed paradigm, we implement a deep learning compiler -- \system. 
Extensive experiments on modern convolution and transformer models show that \system outperforms state-of-the-art DNN inference framework, ONNX Runtime, and compiler, TVM equipped with scheduler AutoTVM and Ansor, by up to \MaxSpeedup (\AvgSpeedup on average).
It also reduces the tuning time by \AutoTVMTuningTimeReduction and \TuningTimeReduction compared with AutoTVM and Ansor, respectively. 
We open-sourced hidet at \url{https://www.github.com/hidet-org/hidet}.

\end{abstract}

\maketitle

\thispagestyle{empty}

\section{Introduction}

Deep neural networks (DNNs)~\cite{lecun2015deep} have achieved state-of-the-art (SOTA) results in various tasks such as image recognition~\cite{he2016deep, krizhevsky2012alexnet,  szegedy2015going, szegedy2016rethinking}, natural language translation~\cite{bert, Lewis2020BARTDS,  sutskever2014seq2seq}, and autonomous driving~\cite{cityscapes}. 
In deployment environments, these models are repeatedly executed to serve continuous user requests, named \emph{model serving}. Thus, it is crucial to reduce the latency and maximize the throughput of model execution to ensure safety, save energy, and improve user experience.

There are two major ways to execute a DNN model.
(1) Deep learning (DL) frameworks such as TensorFlow~\cite{tensorflow}, PyTorch~\cite{pytorch} and ONNX Runtime~\cite{onnxruntime_v1_11_1} dispatch operators to kernel libraries such as cuDNN~\cite{cudnn}, cuBLAS~\cite{cublas}, and CUTLASS~\cite{cutlass} during execution. 
(2) On the other hand, DL compilers such as Tensorflow-XLA~\cite{tensorflow-xla} and TVM~\cite{tvm} automatically generate kernels through a compilation process for the given operators. Various schedulers such as Ansor~\cite{ansor} and AutoTVM~\cite{autotvm} are used to schedule the kernels during compilation to achieve high performance. 

Kernel libraries (e.g., cuDNN~\cite{cudnn} and cuBLAS~\cite{cublas}) provide a collection of highly optimized hand-crafted kernels (e.g., convolutions and matrix multiplications). These libraries typically achieve near-peak performance on widely used input sizes, as 
they are able to implement a large spectrum of optimizations in low-level languages (e.g., CUDA C/C++ and assembly code).
However, manually tweaking a kernel to optimize for performance is laborious, error-prone, and requires expertise in writing low-level language codes.
Thus, it is difficult to generalize to other input shapes, new operators, and kernel fusion patterns.
In addition, template-based libraries such as CUTLASS~\cite{cutlass} employ C++ templates to generate tensor programs for different input shapes on the fly. Although template-based libraries can achieve competitive performance on many input shapes by dynamically tuning the optimization hyper-parameters, they do not reduce the complexity of writing tensor programs for new operators and only provide limited fusion capability (e.g., only a small number of predefined operators can be fused with matrix multiplication).

Alternatively, DL compilers~\cite{\DNNCompilers} are proposed to compile deep learning networks into tensor programs automatically. Existing state-of-the-art DL compilers adopt the idea of decoupling computation definition and scheduling, originally proposed by Halide~\cite{halide} and TVM~\cite{tvm}.
The computation definition of an operator only defines how each element of the output tensor is computed mathematically, and the schedule defines the way the execution is performed, such as the loop order and thread binding~\cite{tvm, halide}.  
Compilers leverage schedulers like AutoTVM~\cite{autotvm} and Asnor~\cite{ansor} to \emph{tune} the hyper-parameters of the schedule to optimize operator performance for each input shape. Unlike kernel libraries and templates that target a fixed set of operators and limited fusion patterns, compilers are capable of supporting more operators and more flexible fusion patterns automatically. 

However, existing state-of-the-art compilers are mostly based on the \emph{loop-oriented scheduling} primitives, which manipulate the loop structure of a tensor program in a \emph{declarative} manner (e.g., loop split and reorder).
Although loop-oriented scheduling primitives have achieved great success in simplifying tensor program writing~\cite{tvm, autotvm, ansor},
certain key optimizations (e.g., double buffering~\cite{cutlass}) are hard to implement.
Specifically, loop-oriented scheduling primitives cannot express the fine-grained tensor program transformations required by the key optimizations discussed in Section~\ref{sec:motivation:challenge1}.
Besides, loop-oriented scheduling also suffers from the long kernel tuning time due to the rarity of efficient schedules in the tremendous
tuning spaces.
For instance, AutoTVM~\cite{autotvm} takes 15 hours to tune a single CNN model Inception V3~\cite{szegedy2016rethinking} on a modern GPU.

In this work, we propose a new paradigm for writing efficient tensor programs: \emph{task-mapping-oriented programming paradigm}. 
In this paradigm, we define the parallelizable computations in an operator as \emph{tasks}, and the process of assigning and ordering the tasks to parallel processing units (e.g., threads) as \emph{scheduling}. The developers can directly define the scheduling in the tensor program through \emph{task mappings}~\footnote{The name task mapping comes from the abstraction where a scheduling process can be considered as the one that maps tasks to processing units in both spatial and temporal dimensions.}.
This paradigm simplifies the development of tensor programs without sacrificing the ability to express optimizations requiring fine-grained program manipulation.
With the in-program style of scheduling, this paradigm also allows us to search the tensor program in an
efficient \emph{hardware-centric schedule space} that is agnostic to input size to dramatically reduce the tuning time.
We also propose \emph{post-scheduling fusion} to fuse the scheduled operator with surrounding operators automatically, so developers don't need to worry about fusion when writing schedule templates.

We implement a new DL compiler called \system
based on the proposed ideas.
In this work, we mainly focus on optimizing DNN inference on GPUs, as it is the most commonly used DNN accelerator. The proposed ideas also apply to other accelerators such as CPUs and TPUs~\cite{tpu}.
Extensive experiments on modern convolutional and transformer models show that \system outperforms state-of-the-art DL inference frameworks and schedulers, AutoTVM~\cite{autotvm} and Ansor~\cite{ansor}, by up to \MaxSpeedup (\AvgSpeedup on average) while reducing the tuning time of the two schedulers by \AutoTVMTuningTimeReduction and \TuningTimeReduction, respectively.

We summarize our contributions as follows:
\begin{itemize}
\item 
We identify and present the limited expressiveness of loop-oriented scheduling adopted by state-of-the-art DL compilers to be their fundamental limitation  in efficiently compiling complex tensor programs (e.g., matrix multiplication).
\item We introduce the task-mapping-oriented programming paradigm to simplify tensor program development without sacrificing the expressiveness of optimizations compared with hand-crafted implementations.
Based on this paradigm, we propose post-scheduling fusion to fuse the scheduled program with surrounding operators.
The paradigm also allows us to search in the hardware-centric schedule space to reduce the tuning time significantly. 
\item We implement a new DL compiler, named \system, based on the proposed ideas. Extensive experiments show that \system outperforms state-of-the-art DL frameworks and compilers by up to \MaxSpeedup and reduces tuning time by \TuningTimeReduction. We have open-sourced Hidet \href{https://github.com/hidet-org/hidet}{here}.
\end{itemize}

\section{Background}

\subsection{CUDA Programming Model}

The CUDA programming platform~\cite{gpu} is widely used by deep learning systems on NVIDIA GPUs. 
In this section, we briefly introduce the CUDA programming model on modern GPUs. 

\begin{figure}[th]%
    \centering
    \includegraphics[width=1.0\linewidth]{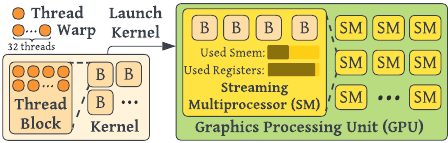} 
    \caption{An overview of CUDA programming model.}%
    \label{fig:background_cuda}%
\end{figure}

\subsubsection*{Kernel, Thread Block, and Thread}
When running a workload on the GPU, thousands of threads will be executed. Each thread executes the same piece of code, called \emph{kernel code}. When launching a kernel, a grid of thread blocks will be dispatched onto the GPU
as shown in Figure~\ref{fig:background_cuda}. Each grid usually comprises tens to thousands of \emph{thread blocks}, while each thread block comprises tens to hundreds of \emph{threads}. In the kernel code, pre-defined variables \code{threadIdx} and \code{blockIdx}, and suffix \code{x}, \code{y}, and \code{z} are used to access the 3-dimensional index of thread in a thread block and the thread block in the grid of blocks, respectively.
\subsubsection*{Hardware Implementation}
Each modern GPU has tens to hundreds of \emph{streaming multiprocessors} (SMs). 
Each SM supports 
scheduling up to thousands of concurrent threads~\cite{gpu}. 
Threads in a thread block are partitioned into \emph{warps}, and each warp contains 32 consecutive threads executing the same instructions.
There are two kinds of programmable on-chip memory: \emph{shared memory} and \emph{registers}. 
Registers are privately allocated to each thread, while
shared memory is allocated to each thread block and only threads in the thread block can access it.
When launching a kernel, the thread blocks are dispatched to the SMs wave by wave~\cite{cta_schedule}. 
Each thread block will only be dispatched to a single SM while each SM may contain multiple thread blocks. 
The number of maximum resident thread blocks per SM is limited by the size of shared memory, register file, and warp scheduling units.

Operators in the deep neural network are implemented as GPU kernels. 
When running a neural network, we launch these kernels following an order satisfying the operator dependency. 
Among these operators, matrix multiplication (also known as a linear or dense layer) is one of the most important operators. 
We next present an efficient implementation of matrix multiplication using CUDA and take it as an example throughout the paper.

\subsection{Efficient Matrix Multiplication}%
\begin{figure}[t]%
    \centering
    \includegraphics[width=1.0\linewidth]{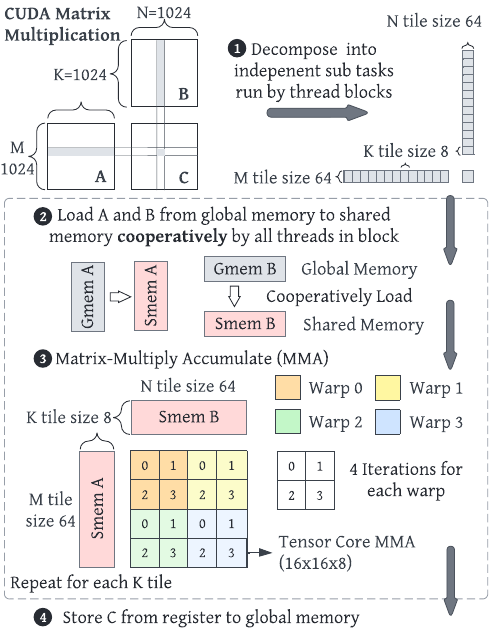} 
    \caption{Efficient Matrix Multiplication on CUDA Platform.}%
    \label{fig:background_matmul}%
\end{figure}
\begin{figure}[th]%
    \centering
    \includegraphics[width=1.0\linewidth]{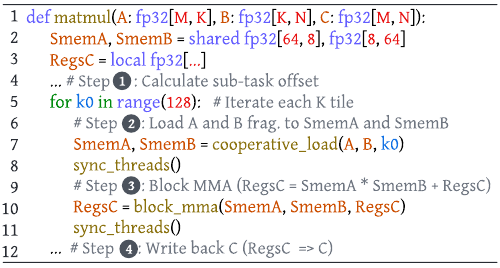} 
    \caption{Pseudo-code of Matrix Multiplication.}%
    \label{fig:background_matmul_code}%
\end{figure}

This section illustrates an efficient implementation of matrix multiplication $C = AB$ (all matrices are $1024\times 1024$) on modern NVIDIA GPUs via Tensor Cores~\cite{a100_tensor_core}.
Figure~\ref{fig:background_matmul} shows the desired workflow. 
In step~\circled{1}, we decompose the matrix multiplication into independent subtasks by tiling the M and N dimensions.
After tiling, there will be $\frac{M}{\text{M tile size}}\times\frac{N}{\text{N tile size}}$ independent subtasks while each sub-task is a matrix multiplication with size: $\text{M tile size} \times \text{N tile size} \times K$.
Each subtask will be assigned to a thread block.
Inside each thread block, the K dimension will be further tiled into $\frac{K}{\text{K tile size}}$ tiles, and the thread block will apply step~\circled{2}-\circled{3} to each K tile.
In step~\circled{2}, threads in the thread block load fragments of matrix A and B from global memory to shared memory collectively (i.e., different threads load different parts of the fragments). All threads in a thread block will be synchronized to make sure the data loading is finished before proceeding to the next step.
In step~\circled{3}, 4 warps in the thread block work on $4\times4=16$ \emph{matrix multiply accumulates} (MMAs), each of which is an operation $C_{16\times16}=A_{16\times8}B_{8\times16}+C_{16\times16}.$
Each warp conducts $4$ MMAs using Tensor Core~\cite{a100_tensor_core} with 4 sequential iterations.
Once we accumulate the results of matrix multiplication for each K tile, we \circled{4} store the results from the accumulating register to global memory.
Figure~\ref{fig:background_matmul_code} gives the pseudo-code of the 4 steps. 

There are two ways to implement the kernel: (1) directly write the CUDA C code as in kernel libraries~\cite{cudnn, cublas, cutlass}, or (2) use declarative loop-oriented scheduling primitives. In the next subsection, we would give a brief introduction to the second method.

\subsection{Declarative Loop-Oriented Scheduling}

To simplify tensor program optimization, Halide~\cite{halide} proposes a programming paradigm of tensor programs, in which the computation definition and scheduling of the computation are decoupled. This programming paradigm is adopted by state-of-the-art DNN compiler TVM~\cite{tvm} and schedulers (e.g., AutoTVM~\cite{autotvm} and Ansor~\cite{ansor}). 
Since this paradigm offers a set of declarative scheduling primitives to manipulate the loop structure of tensor programs, we name it \emph{declarative loop-oriented scheduling}.

\begin{figure}[th]%
    \centering
    \includegraphics[width=0.98\linewidth]{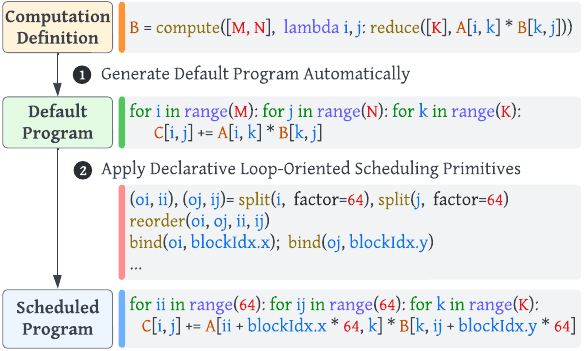} 
    \caption{Workflow of loop-oriented scheduling.}%
    \label{fig:background_loop_oriented_schedule}%
\end{figure}
\begin{table}[ht]
    \centering
    \caption{Loop-oriented scheduling primitives in TVM~\cite{tvm}. 
    The primitive fuse, split, reorder, and bind transforms the program by fusing loop, splitting loop into sub-loops, reordering loops, and binding a loop to a hardware-specific axis.}%
    \vspace{-5pt}
    \includegraphics[width=1.00\linewidth]{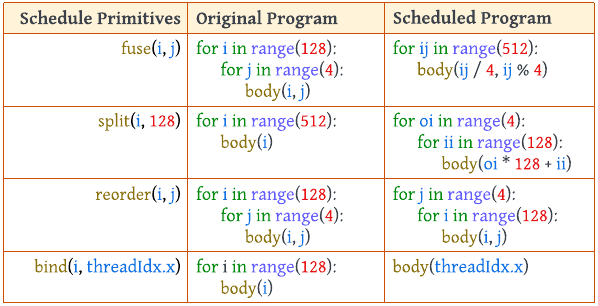} 
    \label{tab:background_primitives}%
    \vspace{-10pt}
\end{table}

Figure~\ref{fig:background_loop_oriented_schedule} shows the workflow of loop-oriented scheduling. 
Developers first provide a mathematical computation of the operator that defines how each element in the tensor is computed. 
The example gives the definition of matrix multiplication, where the (i, j)-th element of the output is a sum reduction. 
Given the computation definition, the schedulers first \circled{1} generate a default tensor program from the computation definition automatically by translating the \code{compute} and \code{reduce} primitives to nested loops. 
Then, a series of \emph{declarative} scheduling primitives are applied to transform the loop structure of the default tensor program for better performance on the specific hardware. 
Table~\ref{tab:background_primitives} shows the scheduling primitives in TVM~\cite{tvm}.\footnote{Schedule primitives that relocate loops are omitted.}
In the example of step \circled{2}, we only list the first few scheduling primitives to implement the matrix multiplication, as TVM has used over 80 primitives to schedule matrix multiplication. 
Starting from the default program, we first split the \code{i} and \code{j} loops with factor 64 into \code{(oi, ii)} and \code{(oj, ij)}, respectively, then reorder loops into \code{(oi, oj, ii, ij)}, and finally bind \code{oi} and \code{oj} to \code{blockIdx.x} and \code{blockIdx.y}, respectively. 
With these primitives, we can get the scheduled program in Figure~\ref{fig:background_loop_oriented_schedule}.

There are several ways to make use of a programming paradigm in a deep learning compiler. 
Intuitively, we can manually write a schedule for each workload (i.e., an operator with a concrete input on certain hardware)~\cite{tvm,halide}. 
However, this approach requires significant engineering efforts to achieve optimal performance for all widely used operators and their typical input sizes.
Consequently, tunable parameters (e.g., tile size and loop orders) are introduced for developers to specify in the schedules. In this way, a manual schedule becomes a \emph{schedule template} and can be optimized by auto-tuning frameworks~\cite{autotvm} for various input shapes and hardware.
To further save the time of writing a schedule template, auto-scheduling approaches that automatically generate a schedule by applying predefined rules to the computation definition have been proposed~\cite{halide_autoscheduler,ansor}.

However, as we illustrate in the next section, 
the schedule space from the loop-oriented scheduling paradigm is still inefficient. 
As a result,
1) it is challenging to achieve competitive performance on operators that are highly optimized by kernel libraries since loop-oriented scheduling can not express some key optimizations,
2) schedulers need hours to days to find the best schedule configuration in the schedule space.

\section{Motivation}

In this section, we summarize the challenges faced by state-of-the-art loop-oriented scheduling.

\subsection{Limited Optimization Support
}
\label{sec:motivation:challenge1}

The declarative loop-oriented scheduling primitives suffer from limited support for key optimizations.
We use an important optimization, \emph{double buffering}~\cite{cuda_dma, cutlass}, 
that has been adopted in several vendor libraries (e.g., cuBLAS~\cite{cublas} and CUTLASS~\cite{cutlass}) but not supported by TVM~\cite{tvm},
to illustrate this fundamental limitation.

The implementation of matrix multiplication in Figure~\ref{fig:background_matmul_code} is sub-optimal since all threads in the same thread blocks are likely to be blocked by one type of hardware resource (i.e., memory bandwidth in Step 2 or computation units in Step 3) while leaving the other idle. 
This is because, in Figure~\ref{fig:background_matmul_code}, the data loading (L7) and computation (L10) use the same buffer, and synchronization (L8) needs to be used to satisfy data dependency.

\begin{figure}[th]%
    \centering
    \includegraphics[width=1.0\linewidth]{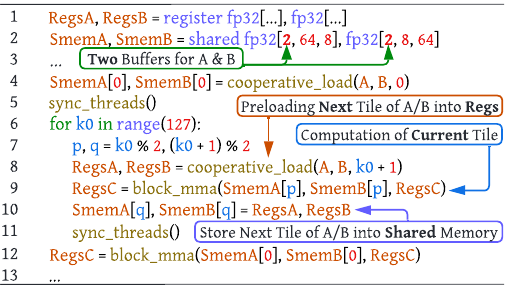} 
    \caption{Double Buffering Optimization.}%
    \label{fig:motivation_matmul_both}%
\end{figure}

The double buffering optimization shown in Figure~\ref{fig:motivation_matmul_both} alleviates the aforementioned problem by using two buffers:
one is used for pre-loading the fragments for the next iteration (\revise{L8 and L10}), while the other is used for computation in the current iteration (\revise{L9}). \revise{We first preload the next tile of matrix A and B into registers (L8), and store them to shared memory after the computation of the current tile (L10). This is more efficient because computation in L9 can be executed while the global memory loading in L8 is on the fly with thread-level parallelism.}
With double buffering, the threads in a thread block can utilize both memory accessing units and computation units at the same time.

However, this optimization cannot be implemented using existing declarative loop-oriented scheduling primitives in Table~\ref{tab:background_primitives}. 
This is because none of the schedule primitives can manipulate the loop body at a fine granularity\footnote{\revise{Even though TVM tried to use a new primitive called \code{double\_buffer} to implement double buffering optimization, it does not separate the global memory loading and shared memory storing, thus can only achieve sub-optimal performance.}}.
As a result, although loop-oriented scheduling simplifies tensor program writing, its \emph{declarative} style of scheduling prevents developers from implementing optimizations requiring fine-grained manipulation of tensor programs.
We want to highlight that double buffering optimization is only an example of the limited expressiveness of existing loop-oriented scheduling.
Besides double buffering, thread block swizzle~\cite{threadblock_swizzle_blog, threadblock_swizzle} and efficient usage\footnote{Directly use MMA PTX instruction instead of WMMA instruction~\cite{ptx}.} of Tensor Core MMA PTX instruction~\cite{ptx}\revise{, and multi-stage asynchronous prefetching~\cite{cutlass}} are widely used optimizations in kernel libraries~\cite{cublas, cutlass}, but are difficult to implement with \revise{declarative} loop-oriented scheduling. To implement these optimizations, we need a more expressive method to write tensor programs and schedule their computations.

\subsection{Dedicated Schedule Template for Fusion}
\label{sec:motivation_fusion}
\begin{figure}[ht]%
    \centering
    \includegraphics[width=0.90\linewidth]{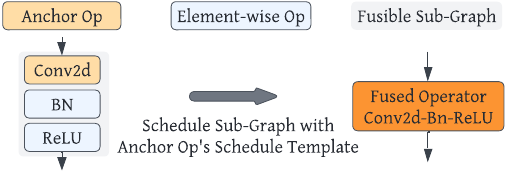} 
    \caption{
    Workflow of TVM sub-graph fusion.
    }%
    \label{fig:motivation_tvm_fusion}%
\end{figure}

One important advantage of compilers over kernel libraries is the ability to optimize arbitrary workloads, especially workloads with multiple fused operators (e.g., \code{Conv2d-BN-ReLU} in convolutional neural networks~\cite{he2016deep}, and \code{Resha\-pe-Matmul-Transpose} in transformer models~\cite{bert}).
For example, Figure~\ref{fig:motivation_tvm_fusion} illustrates how TVM~\cite{tvm} fuses \code{Conv2d-BN-ReLU} into a single kernel.
Specifically, TVM groups operators to form sub-graphs. Each sub-graph can contain only one \emph{anchor} operator, which is usually the most compute-intensive one (e.g., convolution or matrix multiplication) with a carefully designed schedule template.
Then, the schedule template of the anchor operator will be used to schedule the entire sub-graph, meaning that the schedule template has to support all possible fusion scenarios, which greatly increases the complexity of writing schedule templates. 
Although auto-schedulers (e.g., Ansor~\cite{ansor}) are proposed to generate schedule templates automatically from the computation definition
with pre-defined auto-scheduling rules, 
it is challenging to extend the auto-schedulers with new rules. This is because the new rule has to be compatible with all existing rules and needs to be general enough to support all operators.
Thus, it is still challenging to support fusion, while not increasing the complexity of writing specialized schedule templates.

\subsection{Long Tuning Time}
\label{sec:motivation_tuning_time}

\begin{figure}[t]%
    \centering
    \includegraphics[width=1.0\linewidth]{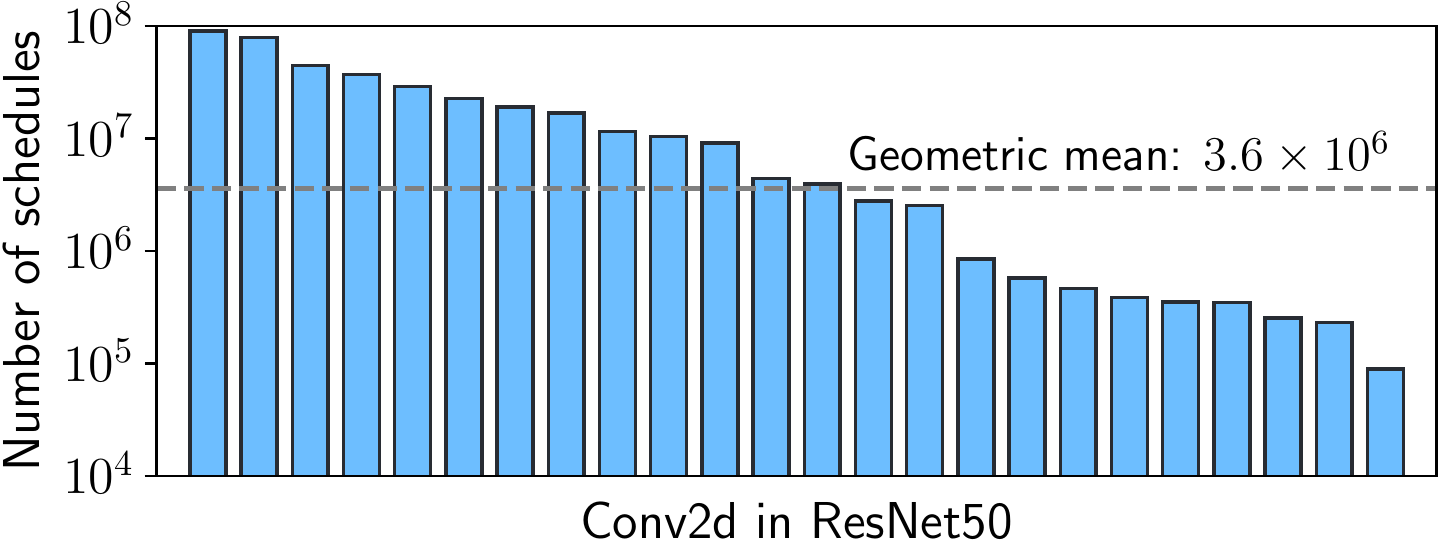} 
    \caption{
    Sizes of schedule spaces adopted by AutoTVM~\cite{autotvm}.
    }%
    \label{fig:motivation_autotvm_schedule_space}%
\end{figure}

In addition to expressiveness and extensibility, the tuning time of existing state-or-the-art schedulers~\cite{halide_autoscheduler, autotvm, ansor} typically ranges from hours to days due to inefficient schedule spaces.
The majority of their schedule spaces are composed of loop tiling factors.
To constrain the schedule space size and avoid conditional if-else branches, 
existing schedulers only cover perfect tile sizes (i.e., only tile $n$-length loop with proper factors of $n$).
For example, potential tile factors of a loop with length $10$ only include $1$, $2$, $5$, and $10$.
As a result, the space constructed by these schedulers with loop-oriented scheduling depends on the input shapes of the target workload. We name this category of schedule space as \emph{input-centric schedule space}. 
We observe two challenges with input-centric schedule space.
(1) The schedule space size grows exponentially along with the number of input size factors. 
Figure~\ref{fig:motivation_autotvm_schedule_space} shows the number of schedules for each convolution in ResNet-50~\cite{he2016deep}. 
There are up to $10^8$ schedules to search for a single convolutional layer.
(2) The schedule space might not include the schedule with optimal performance as non-perfect tile sizes are not considered. An extreme example is that both Ansor and AutoTVM fail to find a valid schedule for matrix multiplication with M=N=K=2039 because 2039 is a prime number.

To address the first challenge, the state-of-the-art schedulers~\cite{autotvm, ansor} employ a cost model to predict the performance of schedules and use genetic evolution search to increase the search efficiency. However, the search process still requires about half an hour to tune a single operator, resulting in 8 to 15 hours to tune an Inception V3 model~\cite{szegedy2016rethinking}. 
Long tuning time prevents existing schedulers from co-optimizing DNNs with graph-level optimizations~\cite{taso, liu2019optimizing} and upper-level applications such as neural architecture search~\cite{rl_nas}. Both of them need the latency of a kernel to guide their optimization and network searching within a short amount of tuning time.

\begin{figure*}[th]%
    \centering
    \includegraphics[width=1.0\linewidth]{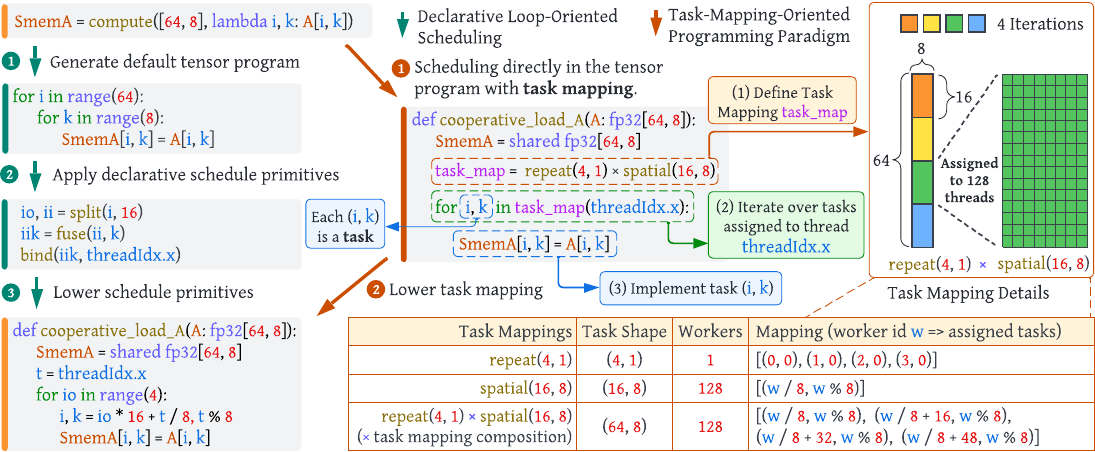} 
    \caption{
    Scheduling the cooperative loading with declarative loop-oriented scheduling and task-mapping programming paradigm. 
    In declarative loop-oriented scheduling, developers apply a series of declarative scheduling primitives to an automatically generated program to transform the tensor program into a more efficient one. Instead of employing declarative primitives, the task-mapping programming paradigm allows developers to directly embed the scheduling in the tensor program and enables a larger spectrum of optimizations compared with loop-oriented scheduling. 
    }%
    \label{fig:key_ideas_task_mapping}%
\end{figure*}

\section{Key Ideas}
\label{sec:key_ideas}

To address the challenges mentioned above,
we propose a new programming paradigm for tensor programs 
-- \emph{task-mapping programming paradigm} (Section~\ref{sec:key_ideas:task_map}). 
This paradigm defines descriptive objects, called \emph{task mapping}, to specify the task assignment and ordering. Task mappings replace the original loop-oriented scheduling primitives and are directly defined and used in the tensor program, which allows more optimizations compared with the existing declarative style of scheduling.
We also propose \emph{post-scheduling fusion} (Section~\ref{sec:key_ideas:fusion}) to simplify sub-graph scheduling by automatically fusing surrounding operators to the operator with \emph{scheduled} tensor program. 
The proposed paradigm also enables efficient partial tiling (tile size is not required to divide loop length) to tune the tensor program in small \emph{hardware-centric schedule space} (Section~\ref{sec:key_ideas:space}) and significantly reduces the tuning time.

\subsection{Task-Mapping Programming Paradigm}
\label{sec:key_ideas:task_map}

Loop-oriented scheduling manipulates a tensor program through \emph{declarative} loop-oriented scheduling primitives to simplify the tensor programming, but at the same time prevents fine-grained manipulations and optimizations. 

We observe that the goal of loop-oriented scheduling primitives is either to (1) assign the computations to parallel processing units (e.g., threads or warps), or (2) specify the execution order of the computations assigned to each processing unit. Figure~\ref{fig:key_ideas_task_mapping} shows the cooperative loading of the matrix A in the matrix multiplication as an example (we omitted the block offset and only show the loading of the matrix A for simplicity). In this example, loop-oriented scheduling applies three primitives (i.e., loop split, fuse, and bind) to assign the loading of 512 (64x8) elements to 128 threads, and each thread loads 4 elements in order.

Instead of scheduling through applying declarative primitives, we propose to embed the scheduling into tensor programs and use dedicated mappings, called \emph{task mappings}, to define the computations 
assignment and ordering directly in the program. 
We use the example in Figure~\ref{fig:key_ideas_task_mapping} to demonstrate how to use task mapping to fulfill the desired scheduling. In \textbf{step (1)}, a task mapping is first defined, which assigns 64x8 tasks to 128 threads. Then, in \textbf{step (2)}, each task \code{(i, k)} assigned to a thread is iterated by calling the task mapping with thread index \code{threadIdx.x}. Finally, in \textbf{step (3)}, the task is implemented using its index \code{(i, k)}.
The three steps decouple the task assignment and the implementation of every single task, greatly simplifying tensor program developments.
Compared with declarative loop-oriented scheduling, it schedules directly in the tensor program and allows more fine-grained optimizations. Besides this, it also allows developers to fall back on some dimensions to traditional loops to implement optimizations such as double buffering~\cite{cutlass}.
Since task mapping is the key component used in the three steps, 
we name our new approach to construct tensor programs -- a \emph{task-mapping programming paradigm}. 

The task mapping defined in step (1) is derived from \emph{task mapping composition} of two basic task mappings (i.e., \code{repeat(4, 1)} and \code{spatial(16, 8)}). 
The table in Figure~\ref{fig:key_ideas_task_mapping} gives the details 
of all appeared task mappings.
The formal definition of task mapping and its composition are given in Section~\ref{sec:tos:task_mapping}.

The proposed paradigm simplifies tensor program development without sacrificing optimization expressiveness. Beyond the scheduling of a single operator, it is also important to schedule a fused sub-graph as operator fusion could greatly reduce the memory traffic to accelerate the end-to-end DNN execution~\cite{tvm, ios, taso}.

\subsection{Post-Scheduling Fusion}
\label{sec:key_ideas:fusion}

\begin{figure}[!ht]%
    \centering
    \includegraphics[width=1.0\linewidth]{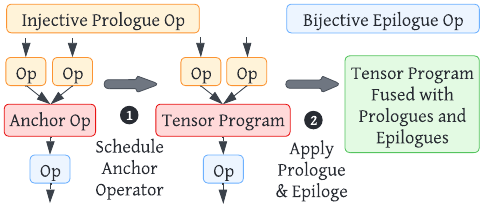}
    \caption{Two steps in post-scheduling fusion.}%
    \label{fig:key_ideas_prologue_epilogue_fusion}%
\end{figure}

We propose to decompose the scheduling of a fused sub-graph into two steps, as shown in Figure~\ref{fig:key_ideas_prologue_epilogue_fusion}.
In step~\circled{1}, we select the anchor operator as TVM~\cite{tvm} does, but only schedule the anchor operator alone. In step~\circled{2}, we fuse the surrounding operators to the \emph{scheduled} tensor program of the anchor operator automatically.  
With this decoupling, the scheduling of the anchor operator does not need to consider the whole sub-graph but only the implementation of itself, which greatly reduces the engineering efforts required to a design schedule template for sub-graph compared with AutoTVM~\cite{autotvm}. Because the fusion is done after we schedule the operator, we call this approach \emph{post-scheduling fusion}.

In post-scheduling fusion, the anchor operator can be fused with operators before (as prologues) and after (as epilogues) it. We decide if an operator is fusible based on its characteristics. If an operator has no reduction computation, it is defined as \emph{injective} and qualified as a prologue operator. If an operator is injective and each element in the input tensor contributes to a single element in the output tensor, it is defined as \emph{bijective} and qualified as an epilogue operator. For example, all elementwise operators (e.g., addition, ReLU~\cite{relu}) and transform operators (e.g., reshape, transpose) are bijective operators and are qualified as both prologue and epilogue operators. 
With post-scheduling fusion, we can concentrate on the scheduling of a single operator while supporting flexible and effective fusion.

\subsection{Hardware-Centric Scheduling Space}
\label{sec:key_ideas:space}

Existing state-of-the-art schedulers~\cite{autotvm, ansor} adopt the input-centric schedule space discussed in Section~\ref{sec:motivation_tuning_time}, in which the schedule chooses the proper factors of loop extent as the split or tile factors, which makes the schedule space unscalable and fails to cover the optimal performance derived from tile sizes that are not proper factors of loop extents.
In addition to constructing a schedule space based on input sizes, another approach is to design the schedule space based on hardware, named \emph{hardware-centric schedule space}. Hardware-centric schedule space decouples the schedule space from the input size by employing predicated loading (i.e., protecting the data loading by checking if the accessing indices are in bounds), and is widely used by kernel libraries~\cite{cudnn, cublas, cutlass}.

With the proposed paradigm, we can provide a small but efficient hardware-centric schedule space.
Since the tile factors are based on hardware resources (e.g., 64x64, 128x64, 16x32, etc), hardware-centric schedule spaces are orders of magnitude smaller than input-centric schedule spaces.
For example, the schedule space we adopted for matrix multiplication contains less than 200 schedules, which is on average $10^5\times$ smaller than a typical schedule space in AutoTVM~\cite{autotvm}. Simply enumerating all schedules would be enough and can be
done within one minute of time.

\section{\system: System Design}

\begin{figure}[ht]%
    \centering
    \includegraphics[width=1.0\linewidth]{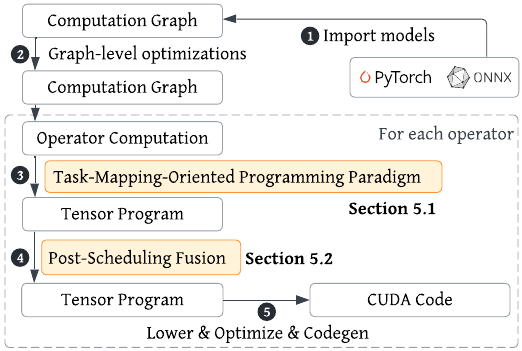} 
    \caption{Overall design of \system.}%
    \label{fig:overall_design}%
\end{figure}

With the above key ideas, we design and implement a DNN 
compiler, named \textbf{\system}.
Figure~\ref{fig:overall_design} shows the overall design.
\system firstly~\circled{1} imports a deep neural network from a widely used framework like PyTorch~\cite{pytorch} or a model file in ONNX~\cite{onnx} format, and then~\circled{2} performs graph-level optimizations, such as constant folding and partition of fusible sub-graphs.
After graph-level optimizations, each anchor operator in the fusible sub-graphs is lowered for scheduling. 
In \system, we \circled{3} schedule the operator with task-mapping programming paradigm (Section~\ref{sec:tos:task_mapping}) into a tensor program and tune the schedule in hardware-centric schedule space.
Then, in step~\circled{4}, the post-scheduling fusion (Section~\ref{sec:tos:fusion}) is applied to fuse the scheduled tensor program of the anchor operator with its surrounding operators automatically.
\circled{5} Finally, the fused tensor programs in Hidet's intermediate representation (IR) will be optimized and lowered. A code generator will convert the lowered IR to CUDA kernels.

\subsection{Task-Mapping Programming Paradigm}
\label{sec:tos:task_mapping}

One key challenge when optimizing tensor programs for certain hardware with parallel processing units (e.g., modern CPUs, GPUs, and TPUs) is how to assign the independent (sub) tasks to the parallel processing units. 
Using cooperative loading in Figure~\ref{fig:key_ideas_task_mapping} as an example, when loading the fragment of matrix A with shape 64x8 from global memory to shared memory, the 512 tasks are assigned to the 128 threads in a thread block, and each thread is assigned with 4 loading tasks.
In this example, tasks are assigned to parallel processing units, called \emph{workers}, and the tasks assigned to each worker will be executed in a specific order. 
In this section, we will first formalize the task assignment and ordering as \emph{task mapping}, then introduce a binary operator on task mappings to compose task mappings, and finally discuss the scheduling based on task mappings.

\subsubsection{Task Mapping}

\begin{figure}[!th]%
    \centering
    \includegraphics[width=0.98\linewidth]{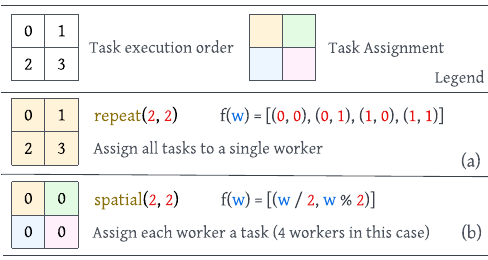} 
    \caption{Examples of two basic kinds of task mappings: \code{repeat(2, 2)} and \code{spatial(2, 2)}. The number indicates the execution order of the tasks assigned to the same task, while the color indicates the worker to which the task was assigned.
    }%
    \label{fig:method_basic_task_mapping}%
\end{figure}

Formally, we define a \emph{worker set} $\mathbf{W}_n$ to be a set containing $n$ workers with id from $0$ to $n-1$: 
$$
\mathbf{W}_n = \{ 0, 1, \dots, n - 1\}.
$$
We also define a task domain $\mathbf{T}$ as
$$
\mathbf{T} = \{(t_0, t_1, \dots, t_{m-1}) \mid 0 \leq t_ i < d_i, t_i \in \mathbb{Z}\},
$$
to represent all tasks we are interested in, where $m$ is the \emph{task dimension} and $\mathbf{d} = (d_0, d_1, \dots, d_{m-1})$ is the \emph{task shape}. 

A \emph{task mapping} $f$ is defined as a function that maps each worker in the worker set to a list of tasks in the task domain, that is
$$
f(w) = [t^{(0)}, t^{(1)}, \dots, t^{(q-1)}].
$$
where $w \in \mathbf{W}$ and $t^{(i)} \in \mathbf{T}$. 

We find two basic task mappings that are very useful. The \code{repeat(d1, ..., dm)} task mapping maps a grid of tasks \code{(d1, ..., dm)} to a single worker while the \code{spatial(d1, ..., dm)} task mapping maps a grid of tasks \code{(d1, ..., dm)} to the same number of workers and each worker only works on a single task. Figure~\ref{fig:method_basic_task_mapping} shows two examples of these task mappings. Besides them, \system also allows developers to define custom task mappings by specifying the task shape, number of workers, and the mapping function. 
Though all examples are in 2-dimension, the task mapping can have an arbitrary number of task dimensions.

\subsubsection{Task Mapping Composition}

\begin{figure}[!th]%
    \centering
    \includegraphics[width=0.98\linewidth]{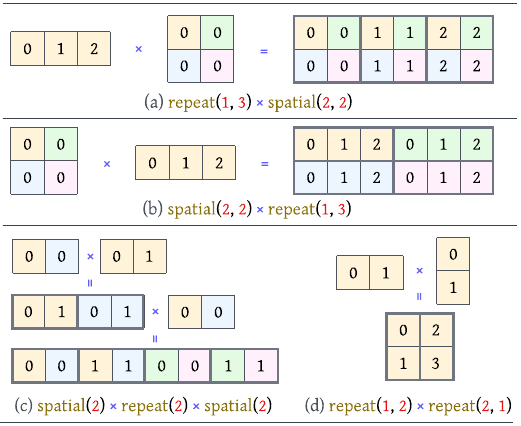} 
    \caption{Examples of task mapping composition. (a) and (b) show that task mapping composition is not communicative. (c) shows an example of composing three task mappings. Because task mapping composition is associative, the order of applying the composition does not matter. (d) shows an example to assign tasks in column-major order.}%
    \label{fig:method_compose_task_mapping}%
\end{figure}

In the example of cooperative loading, we can observe a hierarchical structure. 
The 64x8=512 tasks can be partitioned into 4 groups of tasks and each group contains 16x8=128 tasks. 
The 128 tasks in each group are executed by 128 threads. 
If we take each task group as a macro-task and the 128 threads as a macro-worker, then task-mapping of the macro-tasks to macro-workers is a task mapping that maps 4 tasks to a single worker, denoted by \code{repeat(4, 1)}. 
This example demonstrates that all the tasks in a task mapping can be treated entirely as a single task and all the workers can be treated entirely as a single worker in another task mapping to create a \emph{composed} task mapping.

We formalize this idea as follows. Let $f_1, f_2$ be two task mappings with the same task dimension. Let $n_1, n_2$ be the number of workers and $\mathbf{d_1}, \mathbf{d_2}$ be the task shapes of the two task mappings. We define $f_3$ be the \emph{composed task mapping} of $f_1$ and $f_2$ that has $n_1n_2$ workers and task shape $\mathbf{d_3} = \mathbf{d_1}\odot\mathbf{d_2}$.\footnote{We use $\odot$ to denote the element-wise multiplication.} The mapping function is defined as
$$
f_3(w) = [\mathbf{t_1}\odot\mathbf{d_2}+\mathbf{t_2}\mid \mathbf{t_1} \in f_1(\lfloor w/n_2 \rfloor), \mathbf{t_2} \in f_2(w \;\%\; n_2)].
$$
The task mapping composition is denoted as $f_3 = f_1\circ f_2$. 
Task composition is associative, that is
$$
(f_1\circ f_2)\circ f_3 = f_1\circ(f_2\circ f_3),
$$
holds for arbitrary task mappings $f_1, f_2, f_3$. 

Task mapping composition is a powerful tool to construct new task mappings. Figure~\ref{fig:method_compose_task_mapping} gives some examples of task mapping composition.
Besides these examples, task mapping \code{spatial(4, 2) * repeat(2, 2) * spatial(4, 8) * repeat(4, 4)} is used in matrix multiplication with CUDA Core~\cite{gpu}. 
They correspond to the warps in a block (4x2), the number of repeats for each warp (2x2), the layout of threads in a warp (4x8), and the number of C elements each thread works on (4x4), respectively.

\begin{figure}[th]%
    \centering
    \includegraphics[width=1.0\linewidth]{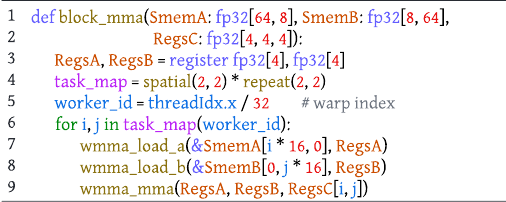} 
    \caption{Use task mapping to schedule warp-level tasks.}%
    \label{fig:method_block_mma}%
\end{figure}

The task and worker in a task mapping are abstract concepts and can be used to describe tasks and workers on different hierarchical levels. For example, besides a single thread, a worker can also represent a warp, a thread block, or a processing unit in other accelerators. Figure~\ref{fig:method_block_mma} shows an example~\footnote{In the example, register buffers \code{RegsA}, \code{RegsB}, and \code{RegsC} are local to each thread. The primitive function \code{wmma\_load\_a} and \code{wmma\_load\_b} load data from shared memory to registers. Primitive function \code{wmma\_mma} conducts the MMA with given registers. The \code{RegsC} has a \emph{special} layout that would map (i, j) to (i\%2, j\%2). For simplicity, we do not introduce the data layouts in \system.} with warps as workers in a task mapping. It implements the \code{block\_mma} function used in the aforementioned matrix multiplication (see Figure~\ref{fig:background_matmul_code} and \ref{fig:motivation_matmul_both}). In the example, we use a task mapping to assign a grid of $4\times4$ tasks to $4$ warps, and each warp takes $4$ warp-level matrix-multiply-accumulate (MMA) task, whose corresponding assignment is shown in step 3 of Figure~\ref{fig:background_matmul}.

Task mappings and their composition could greatly simplify the tensor program writing as it employs dedicated mappings to define the task assignment and ordering, and free developers from writing complex loops and index calculations to achieve the same goal. 
We call the tensor program writing paradigm based on task mappings as \emph{task-mapping programming paradigm} for tensor programs.

\subsubsection{Scheduling Mechanisms}
\label{sec:tos:task_oriented_scheduling}
\begin{figure}[!th]%
    \centering
    \includegraphics[width=1.0\linewidth]{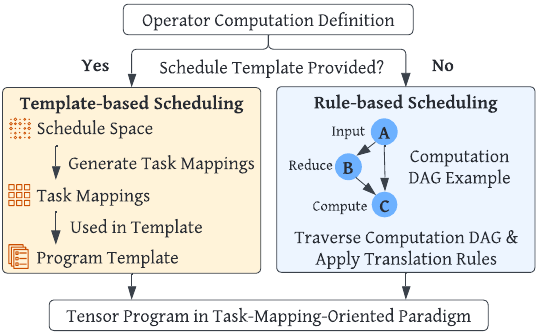} 
    \caption{Two Scheduling Mechanisms in \system.}%
    \label{fig:tos_scheduling}%
\end{figure}

Based on the 
paradigm, 
we further implement two scheduling mechanisms in \system:
\emph{template-based scheduling} and \emph{rule-based scheduling}.
Inspired by Ansor~\cite{ansor} and Halide-AutoScheduler~\cite{halide_autoscheduler}, rule-based scheduling directly generates the tensor program from one operator's computation definition, without any extra engineering efforts and is used for the majority of operators in \system. 
On the other hand, rule-based scheduling might not be able to generate an efficient-enough tensor program for key operators such as matrix multiplication. Inspired by AutoTVM~\cite{autotvm}, we also allow developers to provide a tensor program template to support the efficient scheduling of these operators. Figure~\ref{fig:tos_scheduling} illustrates the two scheduling mechanisms.

\noindent\textbf{Rule-based Scheduling} generates the tensor program given the computation definition automatically. It traverses the computation definition in the form of a directed acyclic graph (DAG) and applies pre-defined rules to translate each node in the DAG into a part of the final tensor program. 
Because this mechanism does not require developers to write a dedicated schedule template, it is widely used in \system for the operators that do not include reduction, such as reshape, transpose, slice, and all element-wise arithmetic operators. 
On the other hand, for operators demanding extreme optimizations like matrix multiplication, we use another scheduling mechanism, named \emph{template-based scheduling}.

\noindent\textbf{Template-based Scheduling} schedules the operator with the given template. A schedule template is a tensor program written with parameterized task mappings. Each schedule template is equipped with a schedule space containing a collection of available parameters for the parameterized task mappings, and the template can be instantiated with an arbitrary choice from the schedule space. 
Taking the matrix multiplication in Figure~\ref{fig:motivation_matmul_both} and ~\ref{fig:method_block_mma} as an example, we could use different numbers of warps and repeat different numbers of times for each warp to implement the matrix multiplication. These different choices form the schedule space for matrix multiplication.
During scheduling, \system first enumerates the schedule choice from the schedule space. Then the schedule choice is used to create the task mappings for the given program template. Finally, \system instantiates the template into a tensor program and measures its performance. The schedule with the best performance is used. We refer to the process as \emph{tuning}. 

\revise{
Adding new operators to Hidet does not require high engineering effort. Most operators in Hidet are scheduled automatically through rule-based scheduling and are easy to add. The computation-intensive operators like convolution and matrix multiplication usually require template-based scheduling for high performance. The complexity of adding a new Hidet template is similar to that of the AutoTVM~\cite{autotvm} template.
}

\subsection{Post-Scheduling Fusion}
\label{sec:tos:fusion}

\begin{figure}[th]%
    \centering
    \includegraphics[width=1.0\linewidth]{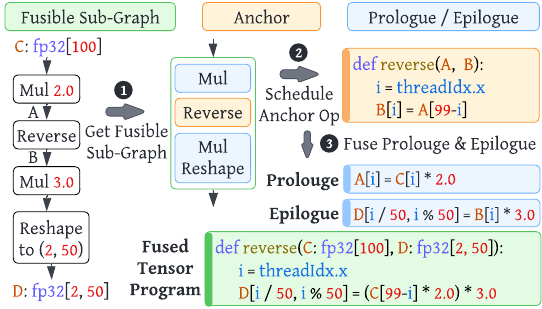} 
    \caption{Example of Post-Scheduling Fusion.}%
    \label{fig:tos_fusion}%
\end{figure}

To alleviate the complexity of scheduling a sub-graph as in AutoTVM, we propose to decouple the sub-graph scheduling into two stages: (1) scheduling the anchor operator and (2) fusing the scheduled tensor program with surrounding operators. The decoupling allows developers to focus on the scheduling of the anchor operator instead of the whole sub-graph, and automates the fusion of the scheduled tensor program with other operators in the sub-graph. During tuning, the performance of fused tensor programs will be used as the target to maximize, thus the decoupling does not hurt the final performance.

Figure~\ref{fig:tos_fusion} shows an example of post-scheduling fusion. 
In step~\circled{1}, during the graph-level optimization stage, an optimization pass partitions the computation graph into sub-graphs. 
Given the sub-graph, in step~\circled{2}, a selected anchor operator will be scheduled into a tensor program with one of the scheduling mechanisms in Section~\ref{sec:tos:task_oriented_scheduling}.
Finally, in step~\circled{3}, the remaining operators will be fused into the scheduled program. 
These operators are classified into two categories: prologue operators for each input tensor and epilogue operators for each output tensor. 
Each prologue operator defines how each element access for the input tensor is computed, and the epilogue operator defines how the output tensor elements are furthermore computed and stored in the output of the fused operator.
In this example, the access of \code{A[99 - i]} will be replaced by \code{C[99 - i] * 2.0}, and the $i$-th element of output tensor is furthermore computed (i.e., multiply by \code{3.0}) and stored to the fused output tensor \code{D} with indices \code{(i / 50, i \% 50)}.

The post-scheduling fusion simplifies the operator fusion. It also allows us to reuse existing highly optimized operators (e.g., matrix multiplication) to support new operators (e.g., convolution). In \system, we can implement the convolution operators as four operators with img2col algorithm~\cite{img2col}, one of which is matrix multiplication and the other three are simple transform operators. With post-scheduling fusion, we fuse the other three operators into a matrix multiplication and reuse all optimizations (e.g., parallel reduction on k dimension~\cite{cutlass}) for matrix multiplications to convolutions.

\begin{figure*}[ht]%
    \centering
    \includegraphics[width=1.00\linewidth]{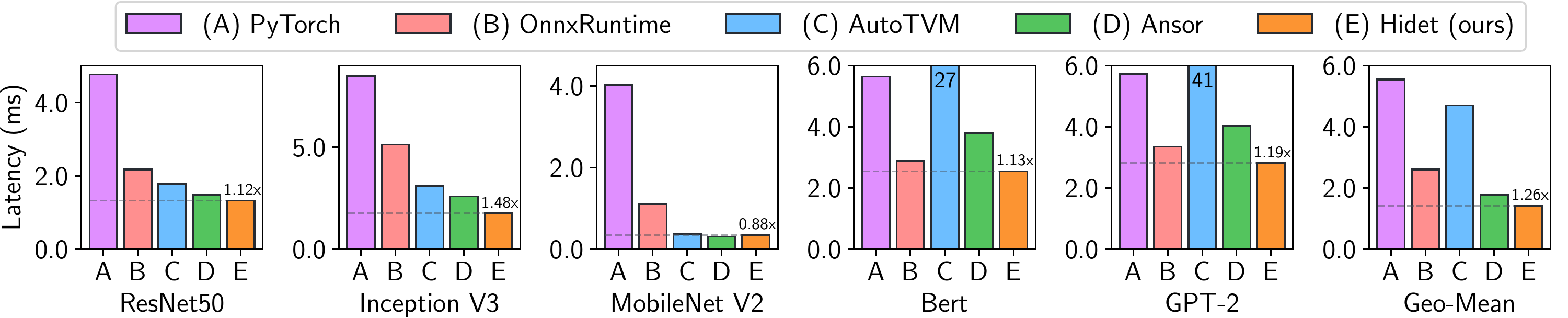} 
    \caption{
    End-to-end comparison between state-of-the-art DNN inference frameworks and compilers with \system.
    }%
    \label{fig:exp_end2end}%
\end{figure*}
\section{Evaluation}
\label{sec:evaluation}

\subsection{Experimental Setup}

\noindent\textbf{Implementation.} We implement \system from scratch with $\sim$20K lines of code in Python and C++. 
Two levels of IR are used in \system: graph-level IR to represent the computation graph of DNN models and tensor-level IR to represent tensor programs with schedules. \system lowers the tensor program written with task mappings to CUDA C code and compiles it with the CUDA compiler.
Notably, we only implement two efficient schedule templates for matrix multiplication and reduction operators (e.g, sum reduction) to cover all operators in evaluated models.
Most operators are either scheduled by the rule-based scheduling mechanism or converted to matrix multiplication to reuse existing templates (e.g., convolutions).

\noindent\textbf{Platform.} We conduct experiments on a server equipped with a 16-core 24-thread Intel i9-12900K CPU (with hyper-threading enabled), 64 GiB DRAM, and one NVIDIA RTX 3090 GPU. The server has installed the Linux distribution Ubuntu LTS 20.04 with NVIDIA driver 510.73.08 and CUDA 11.6. 

\noindent\textbf{Workloads.} We benchmark on a wide range of representative networks to demonstrate the optimization generality of \system. ResNet-50~\cite{he2016deep} is one of the most commonly used CNNs for image classification. Inception-V3~\cite{szegedy2016rethinking} is a CNN that employs multiple paths of convolutions with different kernel sizes.
MobileNet-V2~\cite{mobilenetv2} is a lightweight CNN based on separable convolutions. 
Bert~\cite{bert} is a widely-used transformer-based natural language processing (NLP) model. GPT-2~\cite{gpt2} is an NLP model targeting sequence-to-sequence tasks such as natural language translation and question answering. We use 128 as the sequence length for the two language models throughout the experiments.
We adopt the model implementations in torchvision and transformers packages and export them to ONNX~\cite{onnx} format for evaluation.

\begin{figure}[ht]%
    \centering
    \includegraphics[width=1.00\linewidth]{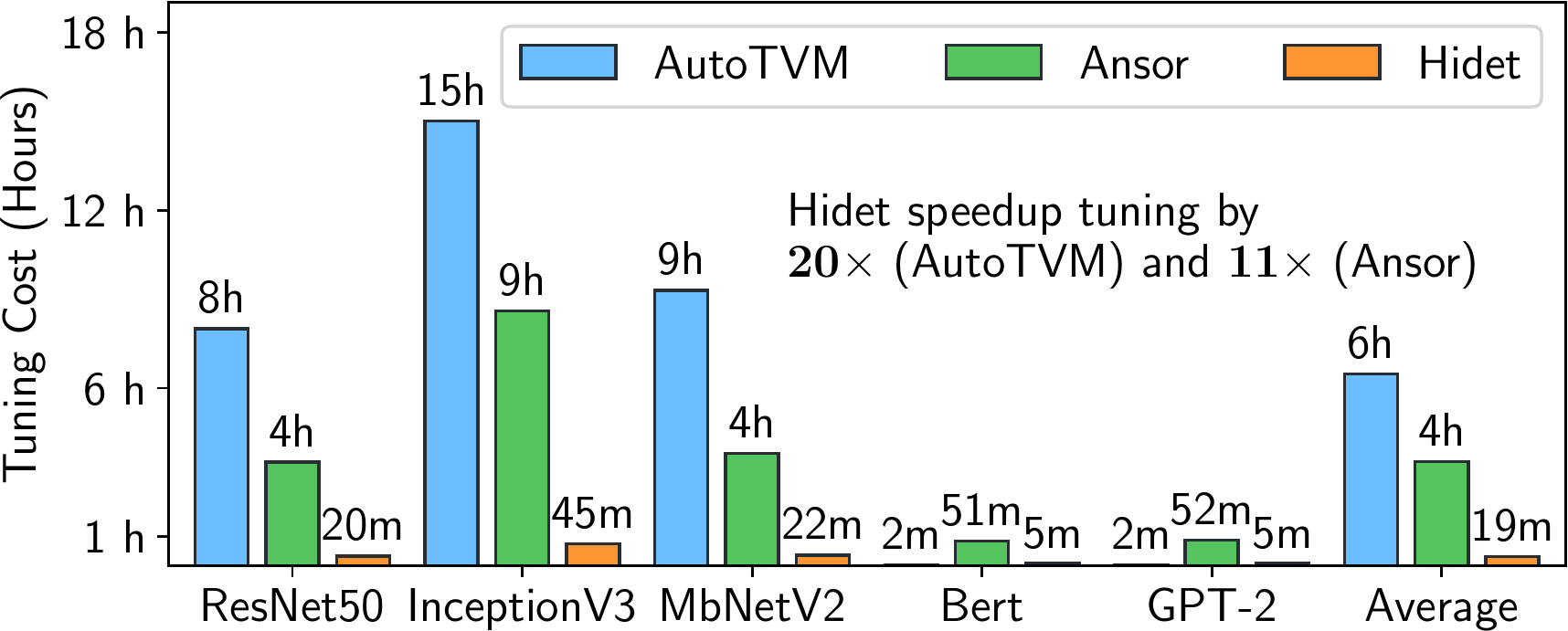} 
    \caption{Tuning cost of AutoTVM, Ansor, and \system.}%
    \label{fig:exp_tuning_time}%
\end{figure}

\subsection{End-to-End Evaluation}
\label{sec:exp:end2end}
\label{sec:exp_tuning_time}

We evaluate all workloads on \system against PyTorch~\cite{pytorch} 1.11, Onnx Runtime~\cite{onnxruntime_v1_11_1} 1.11.1, AutoTVM~\cite{autotvm} and Ansor~\cite{ansor} in TVM~\cite{tvm} 0.9.dev with commit \code{c07a46327}. 
PyTorch is a widely used DNN framework.
Onnx Runtime is a high-performance inference engine. 
Both of them leverage high performance kernel libraries cuDNN~\cite{cudnn} and cuBLAS~\cite{cublas}.
AutoTVM and Ansor are two state-of-the-art schedulers based on loop-oriented scheduling and input-centric tuning spaces.
We set the number of tuning trials in AutoTVM and Ansor to 1000 and 800, respectively, as suggested in their paper and official documentation.

\noindent\textbf{Performance.} Figure~\ref{fig:exp_end2end} shows the results of end-to-end inference latency with a single batch. \system outperforms all baselines on most models by up to \MaxSpeedup, and on average by \AvgSpeedup. This is because \system is able to automatically fuse sub-graph, tune the schedule for given input size (vs. PyTorch and Onnx Runtime), and express more optimizations such as double buffering~\cite{cutlass} (vs. AutoTVM and Ansor).
One exception is Ansor on MobileNetV2, as Ansor could find a better schedule for depthwise convolutions. We can implement similar schedules in \system, and we leave such implementations to future work.
In addition, we note that AutoTVM performs worse on both Bert and GPT-2 models with 27ms and 41ms, respectively. This is because AutoTVM's schedule templates for workloads in these two models lack optimizations.

\noindent\textbf{Tuning Cost.} We compare the tuning cost (i.e., elapsed time in the tuning process) of AutoTVM, Ansor, and \system in Figure~\ref{fig:exp_tuning_time}.
\system reduces the tuning cost by \TuningTimeReduction and \AutoTVMTuningTimeReduction compared with Ansor and AutoTVM, respectively.
This is because \system adopts a small (e.g., $180$ schedules in matrix multiplication) but efficient schedule space with the proposed paradigm. As a result, \system only needs minutes to exhaustively enumerate all candidates.
On the other hand, AutoTVM~\cite{autotvm} and Ansor~\cite{ansor} adopt schedule spaces with $10^5$ to $10^8$ candidates, which prevents them from finding the optimal schedule in their space in a short time, even equipped with a cost model. 
Note that although AutoTVM only spends 2 minutes for Bert and GPT-2 due to their small schedule spaces with less than 20 schedules, the schedule spaces are ineffective and can not achieve competitive performance (Figure~\ref{fig:exp_end2end}).

\subsection{Case Studies}

In this subsection, we conduct several case studies to further demystify the effectiveness of \system.

\subsubsection{Schedule Space Comparison.}
\label{sec:exp:space}

\begin{figure}[th]%
    \centering
    \includegraphics[width=1.0\linewidth]{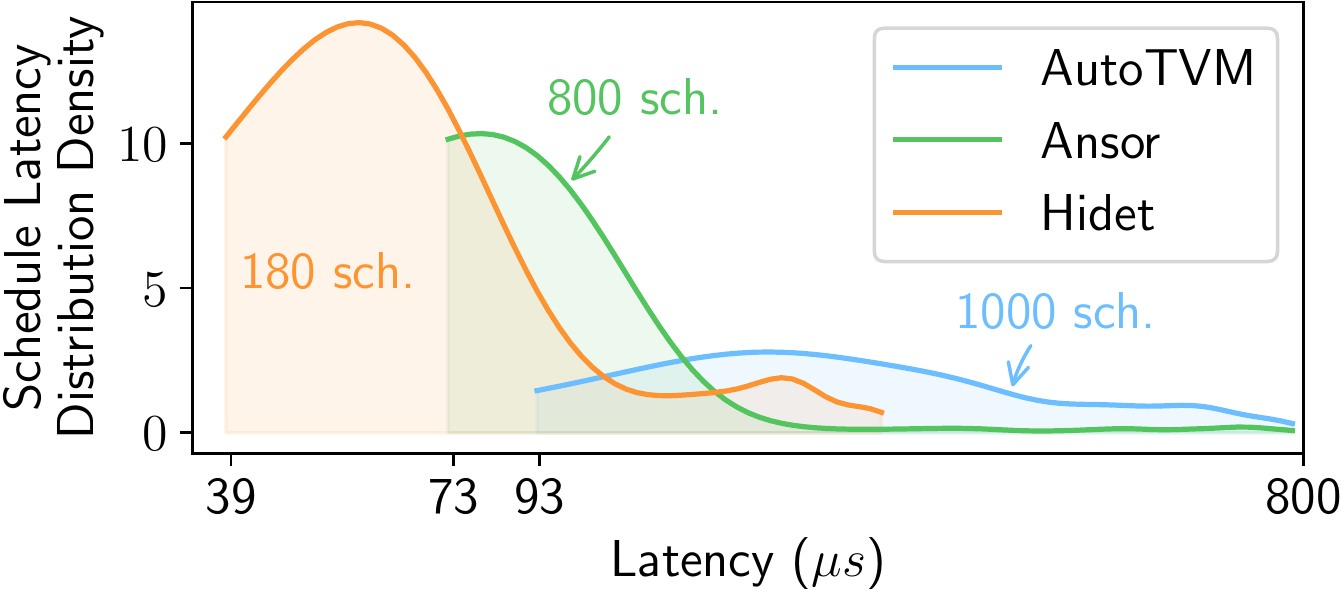} 
    \caption{
    Schedule latency distribution of 
    schedule spaces from AutoTVM, Ansor, and \system. 
    X-axis is in log scale.
    }%
    \label{fig:exp_space_efficiency}%
\end{figure}

To compare the efficiency of three schedule spaces adopted by AutoTVM, Ansor, and \system, we depict the latency distribution of schedules in the three schedule spaces in Figure~\ref{fig:exp_space_efficiency}. 
The benchmark workload is a convolution in ResNet50 with batch size 1, input image size 28x28, input channels 256, kernel size 3, padding 1, and stride 2. 
Because the schedule spaces of AutoTVM and Ansor are too large, we take the 1000 and 800 schedules from the tuning process of AutoTVM and Ansor, respectively, as the samples in their schedule spaces. We compare them with the entire space with only 180 schedules in \system schedule space.
The figure shows that most schedules covered by \system schedule space have superior performance (latency < 73$\mu$s) than those in spaces adopted by AutoTVM and Ansor thanks to the better expressiveness of the proposed paradigm.

\subsubsection{Performance Sensitivity over Input Sizes.}
\label{sec:exp:input_sensitivity}
\begin{figure}[ht]%
    \centering
    \includegraphics[width=1.0\linewidth]{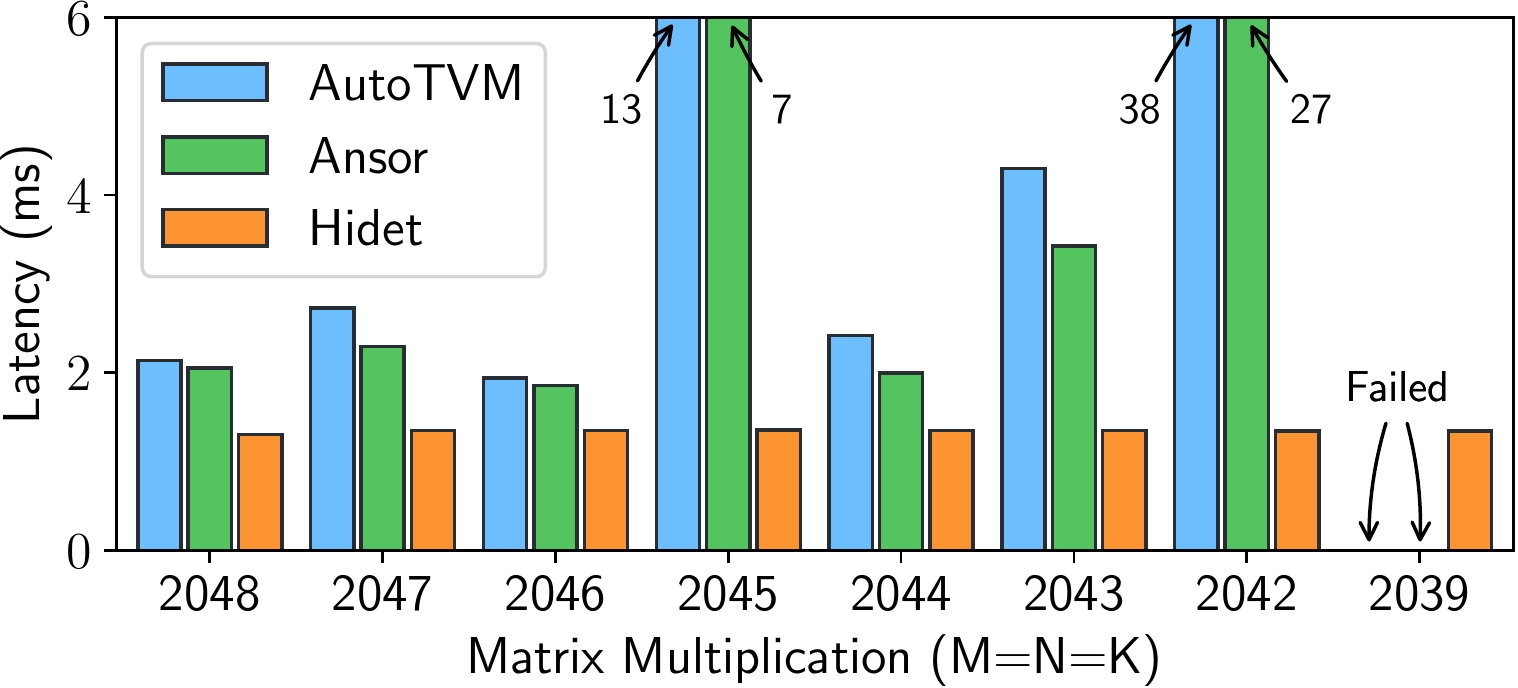} 
    \caption{Comparison of AutoTVM, Ansor, and \system on matrix multiplication with consecutive input sizes. 
    }%
    \label{fig:exp_input_sensitivity}%
\end{figure}
\begin{figure}[ht]%
    \centering
    \includegraphics[width=1.0\linewidth]{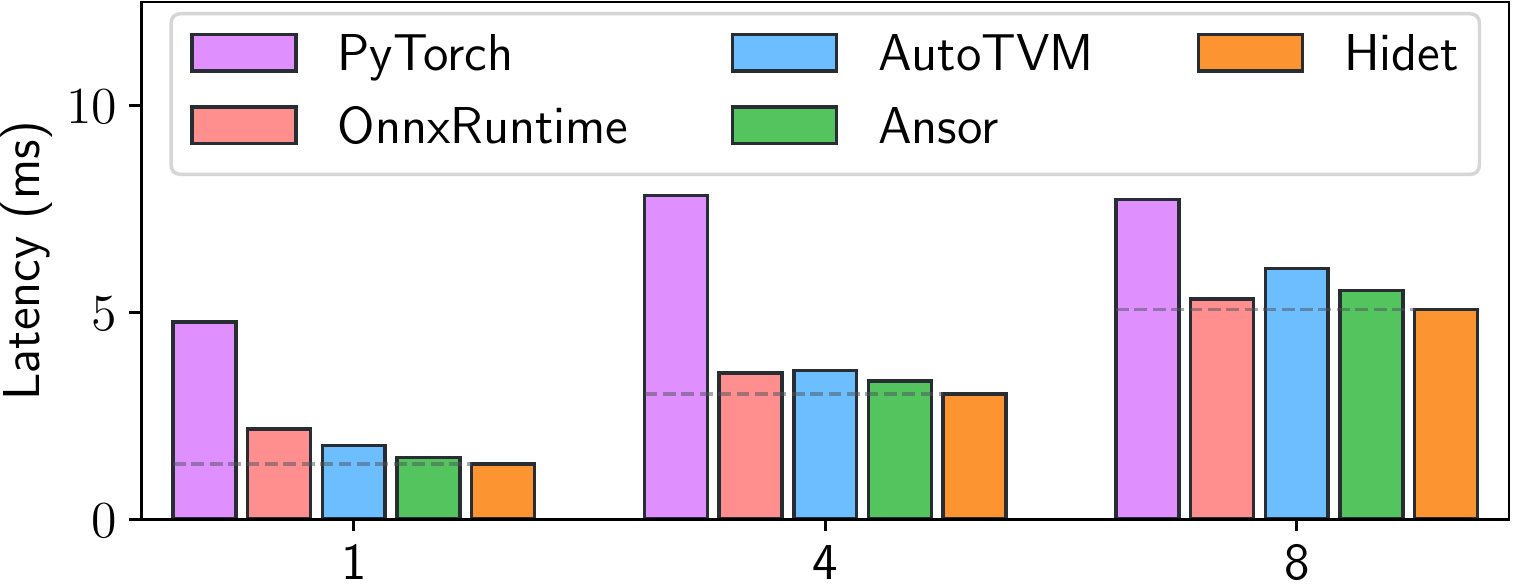} 
    \caption{Comparison on batch size 1, 4, and 8 of ResNet50.}%
    \label{fig:exp_batch_size}%
\end{figure}
\begin{figure}[ht]%
    \centering
    \includegraphics[width=1.0\linewidth]{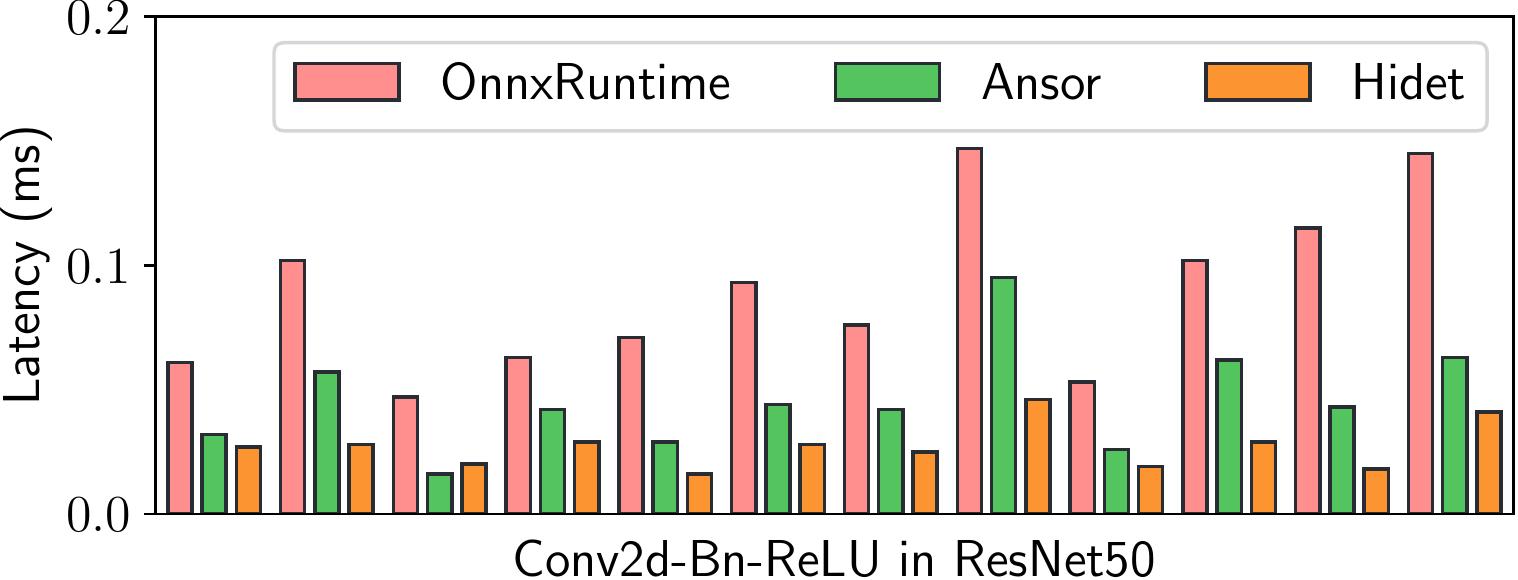} 
    \caption{Comparison of Onnx Runtime, Ansor, and \system on the \code{Conv2d-Bn-ReLU} sub-graphs in ResNet50.}%
    \label{fig:exp_resnet50_conv2ds}%
\end{figure}

The quality of the final schedule derived from AutoTVM and Ansor is sensitive to the input size due to their input-centric schedule spaces.
Even a small change in the input size would result in a large performance difference.
To compare the performance sensitivity over input sizes, we benchmark matrix multiplications with consecutive input sizes. Figure~\ref{fig:exp_input_sensitivity} shows that the performance of AutoTVM and Ansor fluctuates significantly. Even worse, for a prime number input size (e.g., 2039), both schedulers failed to find a valid schedule. On the other hand, with the hardware-centric schedule space, \system achieves consistent performance on these input sizes.

\subsubsection{Evaluation on Different Batch Sizes.}
\label{sec:exp:batch_size}
Figure~\ref{fig:exp_batch_size} depicts the latency of ResNet50 with different batch sizes.
When batch size is small (1 and 4), AutoTVM and Ansor outperform Onnx Runtime as they can find schedules that utilize the GPU computation resources well (e.g., enough thread blocks to saturate all SMs), while kernel libraries do not.
At larger batch sizes (e.g., 8), we observe that although AutoTVM and Ansor can still find schedules that saturate all SMs, they cannot outperform Onnx Runtime, because the latency of each thread block is longer than Onnx Runtime's, due to the lack of important optimizations such as double buffering~\cite{cutlass}.
On the other hand, \system outperforms all of them as \system could perform well on both aspects (i.e., enough and efficient thread blocks).

\subsubsection{Post-Scheduling Fusion Evaluation.}
\label{sec:exp:fusion}

With post-scheduling fusion, we can implement an operator with a highly optimized schedule template, and composite new operators with pre-implemented, highly optimized operators to save engineering efforts. 
For example, in \system, we implement convolution through matrix multiplication, namely implicit general matrix multiplication (GEMM) convolution, which is also known as img2col~\cite{img2col} algorithm. With post-scheduling fusion, we are able to fuse the additional required operators in img2col into the matrix multiplication automatically and reuse the optimizations we implemented for it 
(e.g., parallel reduction on k dimension~\cite{cutlass}). The implicit GEMM convolution with parallel k reduction allows \system's generated kernels to saturate the GPU computation resources and outperforms the existing kernel libraries and DNN compilers. Figure~\ref{fig:exp_resnet50_conv2ds} shows the performance of the \code{Conv-Bn-ReLU} sub-graphs in ResNet50 among Onnx Runtime, Ansor, and \system. \system outperforms Onnx Runtime and Ansor on most convolutions as the convolution can also parallelize on the reduction dimensions (e.g., input channels, and kernel sizes).

\subsubsection{Comparison with TensorRT}
\label{sec:exp:tensorrt}

\begin{figure}[t]%
    \centering
    \includegraphics[width=1.0\linewidth]{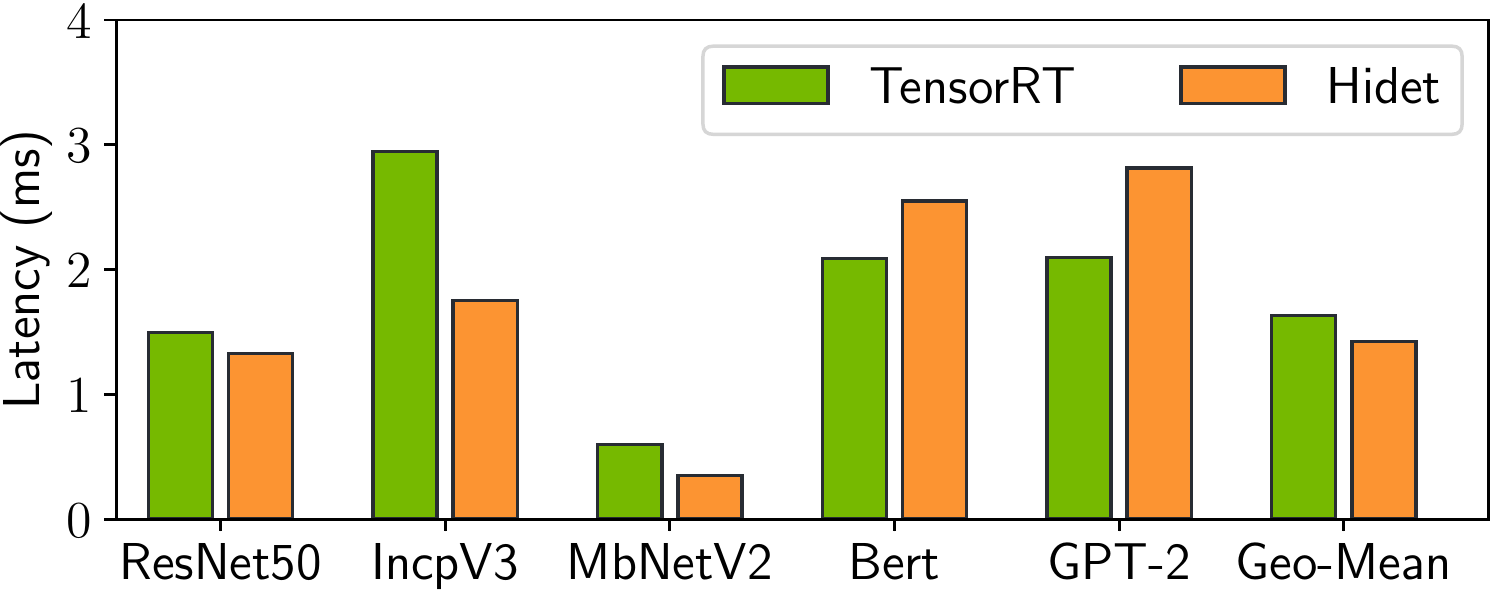} 
    \caption{Comparison of TensorRT and \system. 
    }%
    \vspace{-5pt}
    \label{fig:exp_trt}%
\end{figure}

We also compare \system with TensorRT~\cite{trt} 8.4.1.5, a high-performance deep learning inference engine provided by NVIDIA. 
TensorRT applied both graph-level and operator-level optimizations.
Figure~\ref{fig:exp_trt} shows the comparison of TensorRT and \system. 
\system outperforms TensorRT on the three CNNs because \system is able to tune for the given input sizes and fuse operators automatically according to their mathematical definition. 
On the other hand, TensorRT outperforms \system on the transformer~\cite{attention} networks such as Bert and GPT-2.
Since TensorRT is close-sourced, we speculate, by interpreting its optimization log, that TensorRT recognizes self-attention layers in transformer models and applies dedicated optimizations due to the popularity of these models.
On the other hand, \system only has two schedule templates to cover all operators in benchmarked networks.

\section{Related Work}
\label{sec:related_works}
Many existing DL compilers adopt loop-oriented scheduling primitives \cite{tvm,halide} and establish auto-tuning frameworks on top of them~\cite{halide_autoscheduler,autotvm,tensor_comprehension,unit,bolt,dietcode,ansor,AMOS,flextensor,Shao2022TensorPO} with input-centric schedule spaces.
In contrast, \system leverages task-mapping programming paradigm with hardware-centric schedule spaces, so that it is able to achieve better performance with a much shorter tuning time.
In addition to loop-oriented scheduling, there are more approaches to optimize a tensor program.
Deep learning frameworks such as PyTorch~\cite{pytorch} and TensorFlow~\cite{tensorflow} leverage off-the-shelf kernel libraries (e.g., cuDNN~\cite{cudnn} and cuBLAS~\cite{cublas}) as well as hand-crafted kernels to cover widely used operators.
CUTLASS~\cite{cutlass} is an open C++ template library with efficient matrix multiplication kernels on CUDA. 
Tiramisu~\cite{tiramisu} and AKG~\cite{akg} employ the polyhedral model to schedule the tensor programs.
Roller~\cite{roller} constructs the tensor program with a bottom-up approach and aligns the tile sizes with hardware specifications.
AI-Template~\cite{aitemplate} employs source-code level templates to construct tensor programs, which supports more fine-grained optimizations but sacrifices the flexibility of program transform.
TVM community also noticed the limited expressiveness problem of the existing declarative loop-oriented scheduling mechanism. 
TensorIR~\cite{tensorir}, a concurrent work with Hidet, is recently proposed to allow developers to directly write tensor programs instead of applying a series of declarative primitives to the auto-generated tensor program. 
Moreover, XLA~\cite{tensorflow-xla} is a domain-specific compiler for linear algebra.
FreeTensor~\cite{freetensor} and CoRa~\cite{CoRa} study the compilation for irregular or ragged tensor programs.
AStitch~\cite{astitch} and Apollo~\cite{apollo} study the fusion of memory-intensive kernels to reduce memory consumption.
Fireiron~\cite{fireiron} proposes a data-movement-aware scheduling language for GPUs.
Triton~\cite{triton} proposes to write tensor programs by taking tile as the basic data type and thread block as the main parallel processing unit. 
Nimble~\cite{nimble_haichen}, DISC~\cite{disc}, Cortex~\cite{cortex}, and DietCode~\cite{dietcode} study the compilation of dynamic models, which is also orthogonal with \system. 
Besides optimizing every single operator for DNN inference, Rammer~\cite{rammer} and IOS~\cite{ios} propose to parallelize independent operators in a network. TASO~\cite{taso}, \citet{fang-fusion}, TENSAT~\cite{tensat}, and PET~\cite{pet} apply auto-generated rewriting rules to optimize DNN at the graph level. 
Checkmate~\cite{checkmate}, \citet{chen_remat}, Echo~\cite{echo}, and DTR~\cite{dtr} are proposed to reduce memory footprint.
These works are orthogonal to \system, and can be used to enhance different aspects of \system (e.g., the graph-level optimizations, memory consumption, and dynamic-shape support).

\section{Discussion}
\label{sec:discussion}

\revise{
\textbf{Optimization Expressiveness.} The accelerators (e.g., GPUs and TPUs) usually have a hierarchical memory system and vector- or tensor-based computation engines. Both demand dedicated optimizations to achieve peak performance, and these optimizations are usually hard to be expressed through a series of loop transformations. The double buffering example we discussed in this paper is a good example of such a challenge.
Instead of relying on a declarative style scheduling mechanism, Hidet proposes to directly express the task assignment and ordering with task mapping in a tensor program, which greatly increases the expressiveness of Hidet while reducing the complexity of tensor program writing.
}

\revise{
\textbf{Support More Hardware.} 
Although we only focus on GPUs in this work, the concept of task mapping is general and can be used to describe the task assignment and ordering for other processors. The worker in a task mapping can be 
(1) iterations in a loop for a single-core CPU,
(2) CPU threads for a multi-core CPU, 
(3) threads, warps, or thread blocks for a GPU, and
(4) parallel processing units in other accelerators. And the tasks of a task mapping could be arbitrary indexed, homogeneous, and parallelizable operations.
}

\revise{
\textbf{Future Work.} 
We plan to support CPU and other accelerators (e.g., Amazon Inferentia and Trainium) in the future. 
Besides this, we also plan to support training. Due to the long tuning time of TVM, it is hard to be directly used for accelerating training. Thanks to the hardware-centric schedule space adopted by Hidet, the tuning time has greatly been reduced for Hidet, which makes it possible to build a training system based on Hidet.
}

\section{Conclusion}

We observe that the state-of-the-art DNN compilers based on loop-oriented scheduling cannot express important optimizations that require fine-grained manipulation of the tensor program. 
To address this limitation, we propose task-mapping programming paradigm, a new paradigm to write and schedule tensor programs that simplifies tensor program writing and scheduling without sacrificing the ability to express optimizations as in kernel libraries. 
Based on this paradigm, we implemented a new DNN inference framework called \system. 
Experiments show that \system achieves up to \MaxSpeedup speedup (\AvgSpeedup on average), compared with state-of-the-art DNN  inference  frameworks (e.g., Onnx Runtime) and compilers (e.g., TVM equipped with AutoTVM and Ansor). \system also reduces \TuningTimeReduction tuning cost compared with Ansor.

\balance

\section*{Acknowledgement}

\revise{We would like to thank the members of EcoSystem research laboratory in University of Toronto for their feedback on the early manuscript, and special thanks to Xingyang Song, Christina Giannoula, Anand Jayarajan, and Jiacheng Yang. We also want to thank the anonymous ASPLOS reviewers for the valuable feedback and suggestions, and the artifact evaluation reviewers for reproducing our experiments. The authors with University of Toronto were supported by the Canada Foundation for Innovation JELF grant, NSERC Discovery grant, AWS Machine Learning Research Award (MLRA), Facebook Faculty Research Award, Google Scholar Research Award, and VMware Early Career Faculty Grant.
}

\appendix
\section{Artifact Appendix}

\subsection{Abstract}

This appendix helps readers to reproduce all experiments in the evaluation section via the Hidet artifact~\cite{hidet_artifact}. In Section~\ref{sec:evaluation}, there are 6 experiments (one end to end experiment and 5 case studies). These experiments compare Hidet with other DNN frameworks and compilers on representative DNN models from the perspective of execution latency, optimization time, schedule space, input sensitivity, and different batch sizes. In the public artifact, we provide scripts to launch the 6 experiments automatically. With the hardware and software described in Section~\ref{sec:ae:hardware} and \ref{sec:ae:software}, the artifact should reproduce all experimental results in the evaluation section.

\subsection{Artifact Checklist}

{\small
\begin{itemize}
  \item {\bf Compilation: } NVIDIA CUDA compiler (nvcc).
  \item {\bf Model: } ResNet50, InceptionV3, MobileNetV2, Bert, and GPT-2 
  \item {\bf Run-time environment: } Linux Ubuntu 20.04+
  \item {\bf Hardware: } A workstation equipped with Intel Core i9-12900K, NVIDIA RTX 3090, and 64 GiB RAM.
  \item {\bf Metrics: } End-to-end inference latency and auto-tuning time.
  \item {\bf How much disk space required (approximately)?: } 2 GiB
  \item {\bf How much time is needed to prepare workflow (approximately)?: } 2 hours.
  \item {\bf How much time is needed to complete experiments (approximately)?: } 60 hours. Most of the time (about 50 hours) will be used for model tuning by baselines AutoTVM and Ansor.
  \item {\bf Publicly available?: } Yes
  \item {\bf Code licenses (if publicly available)?: } Apache 2.0.
  \item {\bf Archived (provide DOI)?: } 10.5281/zenodo.7429879

\end{itemize}
}

\subsection{Description}

\subsubsection{How to access}
\label{sec:access}

The source code can be downloaded from either the Zenodo archive ({\small\url{https://doi.org/10.5281/zenodo.7429879}}) or GitHub repository ({\small\url{https://github.com/yaoyaoding/hidet-artifacts}}).

\subsubsection{Hardware dependencies}
\label{sec:ae:hardware}

To get the exact numbers in the evaluation, the exact CPU and GPU is required: Intel Core i9-12900K CPU and NVIDIA RTX 3090 GPU. To functionally run the experiment, the only requirement is a modern NVIDIA GPU that supports CUDA 11.6+.

\subsubsection{Software dependencies}
\label{sec:ae:software}
The artifact requires:
{
\begin{itemize}
    \item NVIDIA Driver 510.73.08
    \item NVIDIA CUDA Toolkit 11.6~\cite{cuda}
    \item NVIDIA kernel library cuDNN 8.4~\cite{cudnn}
    \item Package torch 1.11 (PyTorch~\cite{pytorch})
    \item Package torchvision 0.12 (CNN models~\cite{he2016deep, mobilenetv2, szegedy2016rethinking})
    \item Package transformers 4.19.2 (NLP models~\cite{bert, gpt2})
    \item Package onnxruntime-gpu 1.11.1 (ONNX Runtime~\cite{onnxruntime_v1_11_1})
    \item Package nvidia-tensorrt 8.2.5.1 (Tensor RT~\cite{trt})
    \item Apache TVM 0.9.dev with commit \code{c07a46327} \cite{tvm}
\end{itemize}
}

\subsubsection{Models}

We conduct the experiments with five DNN models: ResNet50~\cite{he2016deep}, InceptionV3~\cite{szegedy2016rethinking}, MobileNetV2~\cite{mobilenetv2}, Bert~\cite{bert}, and GPT-3~\cite{gpt2}. The three convolution networks are from torchvision model zoo, and the two transformer models are from transformers package. All of them will be automatically downloaded.

\subsection{Installation}

Download the source code or clone the git repository in section~\ref{sec:access}. Follow the commands of the installation section in \code{README.md} file under the root of source code directory to build and install hidet and baselines.

\subsection{Experiment Workflow}

There are 6 sub-directories under \code{hidet/artifacts} directory starting with 0, 1, 2, 3, 4, and 5, corresponding the 6 experiments in Section \ref{sec:exp:end2end}, \ref{sec:exp:space}, \ref{sec:exp:input_sensitivity}, \ref{sec:exp:batch_size}, \ref{sec:exp:fusion}, and \ref{sec:exp:tensorrt}. Each sub-directory contains a python script \code{main.py} that can be directly launched to conduct corresponding experiment. 

\subsection{Evaluation and Expected Results}

Each experiment script would have multiple outputs like
{
\small
\begin{verbatim}
BatchSize     Model    Executor    Latency      Std
        1  resnet50       hidet      1.329    0.000
\end{verbatim}
}
that represents the average latency of one executor on a model with a specific batch size in multiple runs. This example shows that it takes Hidet 1.329 ms on average (with standard deviation 0.000 ms) to run a single batch of ResNet50~\cite{he2016deep} model. Some column are omitted here for simplicity.
When conducting the experiments with the hardware and software described in Section~\ref{sec:ae:hardware} and Section~\ref{sec:ae:software}, the artifact should reproduce all experimental results in each evaluation section.

\bibliographystyle{ACM-Reference-Format}
\bibliography{
    references/general.bib, 
    references/system.bib, 
    references/cv.bib, 
    references/nlp.bib
}


\begin{thebibliography}{71}


\ifx \showCODEN    \undefined \def \showCODEN     #1{\unskip}     \fi
\ifx \showDOI      \undefined \def \showDOI       #1{#1}\fi
\ifx \showISBNx    \undefined \def \showISBNx     #1{\unskip}     \fi
\ifx \showISBNxiii \undefined \def \showISBNxiii  #1{\unskip}     \fi
\ifx \showISSN     \undefined \def \showISSN      #1{\unskip}     \fi
\ifx \showLCCN     \undefined \def \showLCCN      #1{\unskip}     \fi
\ifx \shownote     \undefined \def \shownote      #1{#1}          \fi
\ifx \showarticletitle \undefined \def \showarticletitle #1{#1}   \fi
\ifx \showURL      \undefined \def \showURL       {\relax}        \fi
\providecommand\bibfield[2]{#2}
\providecommand\bibinfo[2]{#2}
\providecommand\natexlab[1]{#1}
\providecommand\showeprint[2][]{arXiv:#2}

\bibitem[Abadi et~al\mbox{.}(2015)]%
        {tensorflow}
\bibfield{author}{\bibinfo{person}{Mart\'{i}n Abadi}, \bibinfo{person}{Ashish
  Agarwal}, \bibinfo{person}{Paul Barham}, \bibinfo{person}{Eugene Brevdo},
  \bibinfo{person}{Zhifeng Chen}, \bibinfo{person}{Craig Citro},
  \bibinfo{person}{Greg~S. Corrado}, \bibinfo{person}{Andy Davis},
  \bibinfo{person}{Jeffrey Dean}, \bibinfo{person}{Matthieu Devin},
  \bibinfo{person}{Sanjay Ghemawat}, \bibinfo{person}{Ian Goodfellow},
  \bibinfo{person}{Andrew Harp}, \bibinfo{person}{Geoffrey Irving},
  \bibinfo{person}{Michael Isard}, \bibinfo{person}{Yangqing Jia},
  \bibinfo{person}{Rafal Jozefowicz}, \bibinfo{person}{Lukasz Kaiser},
  \bibinfo{person}{Manjunath Kudlur}, \bibinfo{person}{Josh Levenberg},
  \bibinfo{person}{Dandelion Man\'{e}}, \bibinfo{person}{Rajat Monga},
  \bibinfo{person}{Sherry Moore}, \bibinfo{person}{Derek Murray},
  \bibinfo{person}{Chris Olah}, \bibinfo{person}{Mike Schuster},
  \bibinfo{person}{Jonathon Shlens}, \bibinfo{person}{Benoit Steiner},
  \bibinfo{person}{Ilya Sutskever}, \bibinfo{person}{Kunal Talwar},
  \bibinfo{person}{Paul Tucker}, \bibinfo{person}{Vincent Vanhoucke},
  \bibinfo{person}{Vijay Vasudevan}, \bibinfo{person}{Fernanda Vi\'{e}gas},
  \bibinfo{person}{Oriol Vinyals}, \bibinfo{person}{Pete Warden},
  \bibinfo{person}{Martin Wattenberg}, \bibinfo{person}{Martin Wicke},
  \bibinfo{person}{Yuan Yu}, {and} \bibinfo{person}{Xiaoqiang Zheng}.}
  \bibinfo{year}{2015}\natexlab{}.
\newblock \bibinfo{title}{{TensorFlow}: Large-Scale Machine Learning on
  Heterogeneous Systems}.
\newblock
\newblock
\urldef\tempurl%
\url{https://www.tensorflow.org/}
\showURL{%
\tempurl}
\newblock
\shownote{Software available from tensorflow.org}.


\bibitem[Adams et~al\mbox{.}(2019)]%
        {halide_autoscheduler}
\bibfield{author}{\bibinfo{person}{Andrew Adams}, \bibinfo{person}{Karima Ma},
  \bibinfo{person}{Luke Anderson}, \bibinfo{person}{Riyadh Baghdadi},
  \bibinfo{person}{Tzu-Mao Li}, \bibinfo{person}{Micha\"{e}l Gharbi},
  \bibinfo{person}{Benoit Steiner}, \bibinfo{person}{Steven Johnson},
  \bibinfo{person}{Kayvon Fatahalian}, \bibinfo{person}{Fr\'{e}do Durand},
  {and} \bibinfo{person}{Jonathan Ragan-Kelley}.}
  \bibinfo{year}{2019}\natexlab{}.
\newblock \showarticletitle{Learning to Optimize Halide with Tree Search and
  Random Programs}.
\newblock \bibinfo{journal}{\emph{ACM Trans. Graph.}} \bibinfo{volume}{38},
  \bibinfo{number}{4}, Article \bibinfo{articleno}{121} (\bibinfo{date}{jul}
  \bibinfo{year}{2019}), \bibinfo{numpages}{12}~pages.
\newblock
\showISSN{0730-0301}
\urldef\tempurl%
\url{https://doi.org/10.1145/3306346.3322967}
\showDOI{\tempurl}


\bibitem[Agarap(2018)]%
        {relu}
\bibfield{author}{\bibinfo{person}{Abien~Fred Agarap}.}
  \bibinfo{year}{2018}\natexlab{}.
\newblock \showarticletitle{Deep Learning using Rectified Linear Units (ReLU)}.
\newblock \bibinfo{journal}{\emph{ArXiv}}  \bibinfo{volume}{abs/1803.08375}
  (\bibinfo{year}{2018}).
\newblock


\bibitem[Baghdadi et~al\mbox{.}(2019)]%
        {tiramisu}
\bibfield{author}{\bibinfo{person}{Riyadh Baghdadi}, \bibinfo{person}{Jessica
  Ray}, \bibinfo{person}{Malek~Ben Romdhane}, \bibinfo{person}{Emanuele
  Del~Sozzo}, \bibinfo{person}{Abdurrahman Akkas}, \bibinfo{person}{Yunming
  Zhang}, \bibinfo{person}{Patricia Suriana}, \bibinfo{person}{Shoaib Kamil},
  {and} \bibinfo{person}{Saman Amarasinghe}.} \bibinfo{year}{2019}\natexlab{}.
\newblock \showarticletitle{Tiramisu: A Polyhedral Compiler for Expressing Fast
  and Portable Code}. In \bibinfo{booktitle}{\emph{Proceedings of the 2019
  IEEE/ACM International Symposium on Code Generation and Optimization}}
  (Washington, DC, USA) \emph{(\bibinfo{series}{CGO 2019})}.
  \bibinfo{publisher}{IEEE Press}, \bibinfo{pages}{193–205}.
\newblock
\showISBNx{9781728114361}


\bibitem[Bai et~al\mbox{.}(2019)]%
        {onnx}
\bibfield{author}{\bibinfo{person}{Junjie Bai}, \bibinfo{person}{Fang Lu},
  \bibinfo{person}{Ke Zhang}, {et~al\mbox{.}}} \bibinfo{year}{2019}\natexlab{}.
\newblock \bibinfo{title}{ONNX: Open Neural Network Exchange}.
\newblock \bibinfo{howpublished}{\url{https://github.com/onnx/onnx}}.
\newblock


\bibitem[Bauer et~al\mbox{.}(2011)]%
        {cuda_dma}
\bibfield{author}{\bibinfo{person}{Michael Bauer}, \bibinfo{person}{Henry
  Cook}, {and} \bibinfo{person}{Brucek Khailany}.}
  \bibinfo{year}{2011}\natexlab{}.
\newblock \showarticletitle{CudaDMA: Optimizing GPU memory bandwidth via warp
  specialization}. In \bibinfo{booktitle}{\emph{SC '11: Proceedings of 2011
  International Conference for High Performance Computing, Networking, Storage
  and Analysis}}. \bibinfo{pages}{1--11}.
\newblock
\urldef\tempurl%
\url{https://doi.org/10.1145/2063384.2063400}
\showDOI{\tempurl}


\bibitem[Bavoil(2020)]%
        {threadblock_swizzle_blog}
\bibfield{author}{\bibinfo{person}{Louis Bavoil}.}
  \bibinfo{year}{2020}\natexlab{}.
\newblock \bibinfo{title}{Optimizing Compute Shaders for L2 Locality using
  Thread-Group ID Swizzling}.
\newblock
\newblock
\urldef\tempurl%
\url{https://developer.nvidia.com/blog/optimizing-compute-shaders-for-l2-locality-using-thread-group-id-swizzling/}
\showURL{%
\tempurl}


\bibitem[Chellapilla et~al\mbox{.}(2006)]%
        {img2col}
\bibfield{author}{\bibinfo{person}{Kumar Chellapilla}, \bibinfo{person}{Sidd
  Puri}, {and} \bibinfo{person}{Patrice Simard}.}
  \bibinfo{year}{2006}\natexlab{}.
\newblock \showarticletitle{High performance convolutional neural networks for
  document processing}. In \bibinfo{booktitle}{\emph{Tenth international
  workshop on frontiers in handwriting recognition}}. Suvisoft.
\newblock


\bibitem[Chen et~al\mbox{.}(2018a)]%
        {tvm}
\bibfield{author}{\bibinfo{person}{Tianqi Chen}, \bibinfo{person}{Thierry
  Moreau}, \bibinfo{person}{Ziheng Jiang}, \bibinfo{person}{Lianmin Zheng},
  \bibinfo{person}{Eddie~Q. Yan}, \bibinfo{person}{Haichen Shen},
  \bibinfo{person}{Meghan Cowan}, \bibinfo{person}{Leyuan Wang},
  \bibinfo{person}{Yuwei Hu}, \bibinfo{person}{Luis Ceze},
  \bibinfo{person}{Carlos Guestrin}, {and} \bibinfo{person}{Arvind
  Krishnamurthy}.} \bibinfo{year}{2018}\natexlab{a}.
\newblock \showarticletitle{TVM: An Automated End-to-End Optimizing Compiler
  for Deep Learning}. In \bibinfo{booktitle}{\emph{OSDI}}.
\newblock


\bibitem[Chen et~al\mbox{.}(2016)]%
        {chen_remat}
\bibfield{author}{\bibinfo{person}{Tianqi Chen}, \bibinfo{person}{Bing Xu},
  \bibinfo{person}{Chiyuan Zhang}, {and} \bibinfo{person}{Carlos Guestrin}.}
  \bibinfo{year}{2016}\natexlab{}.
\newblock \bibinfo{title}{Training Deep Nets with Sublinear Memory Cost}.
\newblock
\newblock
\urldef\tempurl%
\url{https://doi.org/10.48550/ARXIV.1604.06174}
\showDOI{\tempurl}


\bibitem[Chen et~al\mbox{.}(2018b)]%
        {autotvm}
\bibfield{author}{\bibinfo{person}{Tianqi Chen}, \bibinfo{person}{Lianmin
  Zheng}, \bibinfo{person}{Eddie Yan}, \bibinfo{person}{Ziheng Jiang},
  \bibinfo{person}{Thierry Moreau}, \bibinfo{person}{Luis Ceze},
  \bibinfo{person}{Carlos Guestrin}, {and} \bibinfo{person}{Arvind
  Krishnamurthy}.} \bibinfo{year}{2018}\natexlab{b}.
\newblock \showarticletitle{Learning to optimize tensor programs}. In
  \bibinfo{booktitle}{\emph{Advances in Neural Information Processing
  Systems}}. \bibinfo{pages}{3389--3400}.
\newblock


\bibitem[Chetlur et~al\mbox{.}(2014)]%
        {cudnn}
\bibfield{author}{\bibinfo{person}{Sharan Chetlur}, \bibinfo{person}{Cliff
  Woolley}, \bibinfo{person}{Philippe Vandermersch},
  \bibinfo{person}{Jonathan~M. Cohen}, \bibinfo{person}{John Tran},
  \bibinfo{person}{Bryan Catanzaro}, {and} \bibinfo{person}{Evan Shelhamer}.}
  \bibinfo{year}{2014}\natexlab{}.
\newblock \showarticletitle{cuDNN: Efficient Primitives for Deep Learning}.
\newblock \bibinfo{journal}{\emph{ArXiv}}  \bibinfo{volume}{abs/1410.0759}
  (\bibinfo{year}{2014}).
\newblock


\bibitem[Choquette et~al\mbox{.}(2021)]%
        {a100_tensor_core}
\bibfield{author}{\bibinfo{person}{Jack Choquette}, \bibinfo{person}{Wishwesh
  Gandhi}, \bibinfo{person}{Olivier Giroux}, \bibinfo{person}{Nick Stam}, {and}
  \bibinfo{person}{Ronny Krashinsky}.} \bibinfo{year}{2021}\natexlab{}.
\newblock \showarticletitle{Nvidia a100 tensor core gpu: Performance and
  innovation}.
\newblock \bibinfo{journal}{\emph{IEEE Micro}} \bibinfo{volume}{41},
  \bibinfo{number}{2} (\bibinfo{year}{2021}), \bibinfo{pages}{29--35}.
\newblock


\bibitem[Cordts et~al\mbox{.}(2016)]%
        {cityscapes}
\bibfield{author}{\bibinfo{person}{Marius Cordts}, \bibinfo{person}{Mohamed
  Omran}, \bibinfo{person}{Sebastian Ramos}, \bibinfo{person}{Timo Rehfeld},
  \bibinfo{person}{Markus Enzweiler}, \bibinfo{person}{Rodrigo Benenson},
  \bibinfo{person}{Uwe Franke}, \bibinfo{person}{Stefan Roth}, {and}
  \bibinfo{person}{Bernt Schiele}.} \bibinfo{year}{2016}\natexlab{}.
\newblock \showarticletitle{The cityscapes dataset for semantic urban scene
  understanding}. In \bibinfo{booktitle}{\emph{Proceedings of the IEEE
  conference on computer vision and pattern recognition}}.
  \bibinfo{pages}{3213--3223}.
\newblock


\bibitem[developers(2021)]%
        {onnxruntime_v1_11_1}
\bibfield{author}{\bibinfo{person}{ONNX~Runtime developers}.}
  \bibinfo{year}{2021}\natexlab{}.
\newblock \bibinfo{title}{ONNX Runtime}.
\newblock \bibinfo{howpublished}{\url{https://onnxruntime.ai/}}.
\newblock
\newblock
\shownote{Version: 1.11.1}.


\bibitem[Devlin et~al\mbox{.}(2018)]%
        {bert}
\bibfield{author}{\bibinfo{person}{Jacob Devlin}, \bibinfo{person}{Ming{-}Wei
  Chang}, \bibinfo{person}{Kenton Lee}, {and} \bibinfo{person}{Kristina
  Toutanova}.} \bibinfo{year}{2018}\natexlab{}.
\newblock \showarticletitle{{BERT:} Pre-training of Deep Bidirectional
  Transformers for Language Understanding}.
\newblock \bibinfo{journal}{\emph{CoRR}}  \bibinfo{volume}{abs/1810.04805}
  (\bibinfo{year}{2018}).
\newblock
\showeprint[arXiv]{1810.04805}
\urldef\tempurl%
\url{http://arxiv.org/abs/1810.04805}
\showURL{%
\tempurl}


\bibitem[Ding(2022)]%
        {hidet_artifact}
\bibfield{author}{\bibinfo{person}{Yaoyao Ding}.}
  \bibinfo{year}{2022}\natexlab{}.
\newblock \bibinfo{booktitle}{\emph{yaoyaoding/hidet-artifacts: DOI Release}}.
\newblock
\urldef\tempurl%
\url{https://doi.org/10.5281/zenodo.7429879}
\showDOI{\tempurl}


\bibitem[Ding et~al\mbox{.}(2021)]%
        {ios}
\bibfield{author}{\bibinfo{person}{Yaoyao Ding}, \bibinfo{person}{Ligeng Zhu},
  \bibinfo{person}{Zhihao Jia}, \bibinfo{person}{Gennady Pekhimenko}, {and}
  \bibinfo{person}{Song Han}.} \bibinfo{year}{2021}\natexlab{}.
\newblock \showarticletitle{Ios: Inter-operator scheduler for cnn
  acceleration}.
\newblock \bibinfo{journal}{\emph{Proceedings of Machine Learning and Systems}}
   \bibinfo{volume}{3} (\bibinfo{year}{2021}), \bibinfo{pages}{167--180}.
\newblock


\bibitem[Fang et~al\mbox{.}(2020)]%
        {fang-fusion}
\bibfield{author}{\bibinfo{person}{Jingzhi Fang}, \bibinfo{person}{Yanyan
  Shen}, \bibinfo{person}{Yue Wang}, {and} \bibinfo{person}{Lei Chen}.}
  \bibinfo{year}{2020}\natexlab{}.
\newblock \showarticletitle{Optimizing DNN Computation Graph Using Graph
  Substitutions}.
\newblock \bibinfo{journal}{\emph{Proc. VLDB Endow.}} \bibinfo{volume}{13},
  \bibinfo{number}{12} (\bibinfo{date}{sep} \bibinfo{year}{2020}),
  \bibinfo{pages}{2734–2746}.
\newblock
\showISSN{2150-8097}
\urldef\tempurl%
\url{https://doi.org/10.14778/3407790.3407857}
\showDOI{\tempurl}


\bibitem[Fegade et~al\mbox{.}(2022)]%
        {CoRa}
\bibfield{author}{\bibinfo{person}{Pratik Fegade}, \bibinfo{person}{Tianqi
  Chen}, \bibinfo{person}{Phillip Gibbons}, {and} \bibinfo{person}{Todd
  Mowry}.} \bibinfo{year}{2022}\natexlab{}.
\newblock \showarticletitle{The CoRa Tensor Compiler: Compilation for Ragged
  Tensors with Minimal Padding}. In \bibinfo{booktitle}{\emph{Proceedings of
  Machine Learning and Systems}},
  \bibfield{editor}{\bibinfo{person}{D.~Marculescu}, \bibinfo{person}{Y.~Chi},
  {and} \bibinfo{person}{C.~Wu}} (Eds.), Vol.~\bibinfo{volume}{4}.
  \bibinfo{pages}{721--747}.
\newblock
\urldef\tempurl%
\url{https://proceedings.mlsys.org/paper/2022/file/d3d9446802a44259755d38e6d163e820-Paper.pdf}
\showURL{%
\tempurl}


\bibitem[Fegade et~al\mbox{.}(2020)]%
        {cortex}
\bibfield{author}{\bibinfo{person}{Pratik Fegade}, \bibinfo{person}{Tianqi
  Chen}, \bibinfo{person}{Phil Gibbons}, {and} \bibinfo{person}{Todd~C.
  Mowry}.} \bibinfo{year}{2020}\natexlab{}.
\newblock \showarticletitle{Cortex: A Compiler for Recursive Deep Learning
  Models}.
\newblock \bibinfo{journal}{\emph{ArXiv}}  \bibinfo{volume}{abs/2011.01383}
  (\bibinfo{year}{2020}).
\newblock


\bibitem[Feng et~al\mbox{.}(2022)]%
        {tensorir}
\bibfield{author}{\bibinfo{person}{Siyuan Feng}, \bibinfo{person}{Bohan Hou},
  \bibinfo{person}{Hongyi Jin}, \bibinfo{person}{Wuwei Lin},
  \bibinfo{person}{Junru Shao}, \bibinfo{person}{Ruihang Lai},
  \bibinfo{person}{Zihao Ye}, \bibinfo{person}{Lianmin Zheng},
  \bibinfo{person}{Cody~Hao Yu}, \bibinfo{person}{Yong Yu}, {and}
  \bibinfo{person}{Tianqi Chen}.} \bibinfo{year}{2022}\natexlab{}.
\newblock \bibinfo{title}{TensorIR: An Abstraction for Automatic Tensorized
  Program Optimization}.
\newblock
\newblock
\urldef\tempurl%
\url{https://doi.org/10.48550/ARXIV.2207.04296}
\showDOI{\tempurl}


\bibitem[Gilman et~al\mbox{.}(2021)]%
        {cta_schedule}
\bibfield{author}{\bibinfo{person}{Guin Gilman}, \bibinfo{person}{Samuel~S.
  Ogden}, \bibinfo{person}{Tian Guo}, {and} \bibinfo{person}{Robert~J. Walls}.}
  \bibinfo{year}{2021}\natexlab{}.
\newblock \showarticletitle{Demystifying the Placement Policies of the NVIDIA
  GPU Thread Block Scheduler for Concurrent Kernels}.
\newblock \bibinfo{journal}{\emph{SIGMETRICS Perform. Eval. Rev.}}
  \bibinfo{volume}{48}, \bibinfo{number}{3} (\bibinfo{date}{mar}
  \bibinfo{year}{2021}), \bibinfo{pages}{81–88}.
\newblock
\showISSN{0163-5999}
\urldef\tempurl%
\url{https://doi.org/10.1145/3453953.3453972}
\showDOI{\tempurl}


\bibitem[Hagedorn et~al\mbox{.}(2020)]%
        {fireiron}
\bibfield{author}{\bibinfo{person}{Bastian Hagedorn},
  \bibinfo{person}{Archibald~Samuel Elliott}, \bibinfo{person}{Henrik
  Barthels}, \bibinfo{person}{Rastislav Bodik}, {and} \bibinfo{person}{Vinod
  Grover}.} \bibinfo{year}{2020}\natexlab{}.
\newblock \showarticletitle{Fireiron: A Data-Movement-Aware Scheduling Language
  for GPUs}. In \bibinfo{booktitle}{\emph{Proceedings of the ACM International
  Conference on Parallel Architectures and Compilation Techniques}} (Virtual
  Event, GA, USA) \emph{(\bibinfo{series}{PACT '20})}.
  \bibinfo{publisher}{Association for Computing Machinery},
  \bibinfo{address}{New York, NY, USA}, \bibinfo{pages}{71–82}.
\newblock
\showISBNx{9781450380751}
\urldef\tempurl%
\url{https://doi.org/10.1145/3410463.3414632}
\showDOI{\tempurl}


\bibitem[He et~al\mbox{.}(2016)]%
        {he2016deep}
\bibfield{author}{\bibinfo{person}{Kaiming He}, \bibinfo{person}{Xiangyu
  Zhang}, \bibinfo{person}{Shaoqing Ren}, {and} \bibinfo{person}{Jian Sun}.}
  \bibinfo{year}{2016}\natexlab{}.
\newblock \showarticletitle{Deep residual learning for image recognition}. In
  \bibinfo{booktitle}{\emph{Proceedings of the IEEE conference on computer
  vision and pattern recognition}}. \bibinfo{pages}{770--778}.
\newblock


\bibitem[Inc.(2022a)]%
        {cublas}
\bibfield{author}{\bibinfo{person}{NVIDIA Inc.}}
  \bibinfo{year}{2022}\natexlab{a}.
\newblock \bibinfo{title}{Basic Linear Algebra on NVIDIA GPUs}.
\newblock
\newblock
\urldef\tempurl%
\url{https://developer.nvidia.com/cublas}
\showURL{%
\tempurl}


\bibitem[Inc.(2022b)]%
        {trt}
\bibfield{author}{\bibinfo{person}{NVIDIA Inc.}}
  \bibinfo{year}{2022}\natexlab{b}.
\newblock \bibinfo{title}{NVIDIA TensorRT}.
\newblock
\newblock
\urldef\tempurl%
\url{https://developer.nvidia.com/tensorrt}
\showURL{%
\tempurl}


\bibitem[Inc.(2022c)]%
        {ptx}
\bibfield{author}{\bibinfo{person}{NVIDIA Inc.}}
  \bibinfo{year}{2022}\natexlab{c}.
\newblock \bibinfo{title}{Parallel Thread Execution ISA}.
\newblock
\newblock
\urldef\tempurl%
\url{https://docs.nvidia.com/cuda/parallel-thread-execution/index.html}
\showURL{%
\tempurl}


\bibitem[Jain et~al\mbox{.}(2020)]%
        {checkmate}
\bibfield{author}{\bibinfo{person}{Paras Jain}, \bibinfo{person}{Ajay Jain},
  \bibinfo{person}{Aniruddha Nrusimha}, \bibinfo{person}{Amir Gholami},
  \bibinfo{person}{Pieter Abbeel}, \bibinfo{person}{Joseph Gonzalez},
  \bibinfo{person}{Kurt Keutzer}, {and} \bibinfo{person}{Ion Stoica}.}
  \bibinfo{year}{2020}\natexlab{}.
\newblock \showarticletitle{Checkmate: Breaking the Memory Wall with Optimal
  Tensor Rematerialization}. In \bibinfo{booktitle}{\emph{Proceedings of
  Machine Learning and Systems}},
  \bibfield{editor}{\bibinfo{person}{I.~Dhillon},
  \bibinfo{person}{D.~Papailiopoulos}, {and} \bibinfo{person}{V.~Sze}} (Eds.),
  Vol.~\bibinfo{volume}{2}. \bibinfo{pages}{497--511}.
\newblock
\urldef\tempurl%
\url{https://proceedings.mlsys.org/paper/2020/file/084b6fbb10729ed4da8c3d3f5a3ae7c9-Paper.pdf}
\showURL{%
\tempurl}


\bibitem[Jia et~al\mbox{.}(2019)]%
        {taso}
\bibfield{author}{\bibinfo{person}{Zhihao Jia}, \bibinfo{person}{Oded Padon},
  \bibinfo{person}{James Thomas}, \bibinfo{person}{Todd Warszawski},
  \bibinfo{person}{Matei Zaharia}, {and} \bibinfo{person}{Alex Aiken}.}
  \bibinfo{year}{2019}\natexlab{}.
\newblock \showarticletitle{TASO: optimizing deep learning computation with
  automatic generation of graph substitutions}. In
  \bibinfo{booktitle}{\emph{Proceedings of the 27th ACM Symposium on Operating
  Systems Principles}}. \bibinfo{pages}{47--62}.
\newblock


\bibitem[Jouppi et~al\mbox{.}(2017)]%
        {tpu}
\bibfield{author}{\bibinfo{person}{Norman~P. Jouppi}, \bibinfo{person}{Cliff
  Young}, \bibinfo{person}{Nishant Patil}, \bibinfo{person}{David Patterson},
  \bibinfo{person}{Gaurav Agrawal}, \bibinfo{person}{Raminder Bajwa},
  \bibinfo{person}{Sarah Bates}, \bibinfo{person}{Suresh Bhatia},
  \bibinfo{person}{Nan Boden}, \bibinfo{person}{Al Borchers},
  \bibinfo{person}{Rick Boyle}, \bibinfo{person}{Pierre-luc Cantin},
  \bibinfo{person}{Clifford Chao}, \bibinfo{person}{Chris Clark},
  \bibinfo{person}{Jeremy Coriell}, \bibinfo{person}{Mike Daley},
  \bibinfo{person}{Matt Dau}, \bibinfo{person}{Jeffrey Dean},
  \bibinfo{person}{Ben Gelb}, \bibinfo{person}{Tara~Vazir Ghaemmaghami},
  \bibinfo{person}{Rajendra Gottipati}, \bibinfo{person}{William Gulland},
  \bibinfo{person}{Robert Hagmann}, \bibinfo{person}{C.~Richard Ho},
  \bibinfo{person}{Doug Hogberg}, \bibinfo{person}{John Hu},
  \bibinfo{person}{Robert Hundt}, \bibinfo{person}{Dan Hurt},
  \bibinfo{person}{Julian Ibarz}, \bibinfo{person}{Aaron Jaffey},
  \bibinfo{person}{Alek Jaworski}, \bibinfo{person}{Alexander Kaplan},
  \bibinfo{person}{Harshit Khaitan}, \bibinfo{person}{Daniel Killebrew},
  \bibinfo{person}{Andy Koch}, \bibinfo{person}{Naveen Kumar},
  \bibinfo{person}{Steve Lacy}, \bibinfo{person}{James Laudon},
  \bibinfo{person}{James Law}, \bibinfo{person}{Diemthu Le},
  \bibinfo{person}{Chris Leary}, \bibinfo{person}{Zhuyuan Liu},
  \bibinfo{person}{Kyle Lucke}, \bibinfo{person}{Alan Lundin},
  \bibinfo{person}{Gordon MacKean}, \bibinfo{person}{Adriana Maggiore},
  \bibinfo{person}{Maire Mahony}, \bibinfo{person}{Kieran Miller},
  \bibinfo{person}{Rahul Nagarajan}, \bibinfo{person}{Ravi Narayanaswami},
  \bibinfo{person}{Ray Ni}, \bibinfo{person}{Kathy Nix},
  \bibinfo{person}{Thomas Norrie}, \bibinfo{person}{Mark Omernick},
  \bibinfo{person}{Narayana Penukonda}, \bibinfo{person}{Andy Phelps},
  \bibinfo{person}{Jonathan Ross}, \bibinfo{person}{Matt Ross},
  \bibinfo{person}{Amir Salek}, \bibinfo{person}{Emad Samadiani},
  \bibinfo{person}{Chris Severn}, \bibinfo{person}{Gregory Sizikov},
  \bibinfo{person}{Matthew Snelham}, \bibinfo{person}{Jed Souter},
  \bibinfo{person}{Dan Steinberg}, \bibinfo{person}{Andy Swing},
  \bibinfo{person}{Mercedes Tan}, \bibinfo{person}{Gregory Thorson},
  \bibinfo{person}{Bo Tian}, \bibinfo{person}{Horia Toma},
  \bibinfo{person}{Erick Tuttle}, \bibinfo{person}{Vijay Vasudevan},
  \bibinfo{person}{Richard Walter}, \bibinfo{person}{Walter Wang},
  \bibinfo{person}{Eric Wilcox}, {and} \bibinfo{person}{Doe~Hyun Yoon}.}
  \bibinfo{year}{2017}\natexlab{}.
\newblock \showarticletitle{In-Datacenter Performance Analysis of a Tensor
  Processing Unit}.
\newblock \bibinfo{journal}{\emph{SIGARCH Comput. Archit. News}}
  \bibinfo{volume}{45}, \bibinfo{number}{2} (\bibinfo{date}{jun}
  \bibinfo{year}{2017}), \bibinfo{pages}{1–12}.
\newblock
\showISSN{0163-5964}
\urldef\tempurl%
\url{https://doi.org/10.1145/3140659.3080246}
\showDOI{\tempurl}


\bibitem[Kerr et~al\mbox{.}(2018)]%
        {cutlass}
\bibfield{author}{\bibinfo{person}{Andrew Kerr}, \bibinfo{person}{Duane
  Merrill}, \bibinfo{person}{Julien Demouth}, \bibinfo{person}{John Tran},
  \bibinfo{person}{Naila Farooqui}, \bibinfo{person}{Markus Tavenrath},
  \bibinfo{person}{Vince Schuster}, \bibinfo{person}{Eddie Gornish},
  \bibinfo{person}{Jerry Zheng}, {and} \bibinfo{person}{Bageshri Sathe}.}
  \bibinfo{year}{2018}\natexlab{}.
\newblock \bibinfo{title}{CUTLASS: CUDA TEMPLATE LIBRARY FOR DENSE LINEAR
  ALGEBRA AT ALL LEVELS AND SCALES}.
\newblock
\newblock


\bibitem[Kirisame et~al\mbox{.}(2021)]%
        {dtr}
\bibfield{author}{\bibinfo{person}{Marisa Kirisame}, \bibinfo{person}{Steven
  Lyubomirsky}, \bibinfo{person}{Altan Haan}, \bibinfo{person}{Jennifer
  Brennan}, \bibinfo{person}{Mike He}, \bibinfo{person}{Jared Roesch},
  \bibinfo{person}{Tianqi Chen}, {and} \bibinfo{person}{Zachary Tatlock}.}
  \bibinfo{year}{2021}\natexlab{}.
\newblock \showarticletitle{Dynamic Tensor Rematerialization}. In
  \bibinfo{booktitle}{\emph{9th International Conference on Learning
  Representations, {ICLR} 2021, Virtual Event, Austria, May 3-7, 2021}}.
  \bibinfo{publisher}{OpenReview.net}.
\newblock
\urldef\tempurl%
\url{https://openreview.net/forum?id=Vfs\_2RnOD0H}
\showURL{%
\tempurl}


\bibitem[Krizhevsky et~al\mbox{.}(2012)]%
        {krizhevsky2012alexnet}
\bibfield{author}{\bibinfo{person}{Alex Krizhevsky}, \bibinfo{person}{Ilya
  Sutskever}, {and} \bibinfo{person}{Geoffrey~E Hinton}.}
  \bibinfo{year}{2012}\natexlab{}.
\newblock \showarticletitle{Imagenet classification with deep convolutional
  neural networks}. In \bibinfo{booktitle}{\emph{Advances in neural information
  processing systems}}. \bibinfo{pages}{1097--1105}.
\newblock


\bibitem[LeCun et~al\mbox{.}(2015)]%
        {lecun2015deep}
\bibfield{author}{\bibinfo{person}{Yann LeCun}, \bibinfo{person}{Yoshua
  Bengio}, {and} \bibinfo{person}{Geoffrey Hinton}.}
  \bibinfo{year}{2015}\natexlab{}.
\newblock \showarticletitle{Deep learning}.
\newblock \bibinfo{journal}{\emph{nature}} \bibinfo{volume}{521},
  \bibinfo{number}{7553} (\bibinfo{year}{2015}), \bibinfo{pages}{436--444}.
\newblock


\bibitem[Lewis et~al\mbox{.}(2020)]%
        {Lewis2020BARTDS}
\bibfield{author}{\bibinfo{person}{Mike Lewis}, \bibinfo{person}{Yinhan Liu},
  \bibinfo{person}{Naman Goyal}, \bibinfo{person}{Marjan Ghazvininejad},
  \bibinfo{person}{Abdelrahman Mohamed}, \bibinfo{person}{Omer Levy},
  \bibinfo{person}{Veselin Stoyanov}, {and} \bibinfo{person}{Luke
  Zettlemoyer}.} \bibinfo{year}{2020}\natexlab{}.
\newblock \showarticletitle{BART: Denoising Sequence-to-Sequence Pre-training
  for Natural Language Generation, Translation, and Comprehension}. In
  \bibinfo{booktitle}{\emph{ACL}}.
\newblock


\bibitem[Liu et~al\mbox{.}(2019)]%
        {liu2019optimizing}
\bibfield{author}{\bibinfo{person}{Yizhi Liu}, \bibinfo{person}{Yao Wang},
  \bibinfo{person}{Ruofei Yu}, \bibinfo{person}{Mu Li}, \bibinfo{person}{Vin
  Sharma}, {and} \bibinfo{person}{Yida Wang}.} \bibinfo{year}{2019}\natexlab{}.
\newblock \showarticletitle{Optimizing $\{$CNN$\}$ Model Inference on
  $\{$CPUs$\}$}. In \bibinfo{booktitle}{\emph{2019 USENIX Annual Technical
  Conference (USENIX ATC 19)}}. \bibinfo{pages}{1025--1040}.
\newblock


\bibitem[Ma et~al\mbox{.}(2020)]%
        {rammer}
\bibfield{author}{\bibinfo{person}{Lingxiao Ma}, \bibinfo{person}{Zhiqiang
  Xie}, \bibinfo{person}{Zhi Yang}, \bibinfo{person}{Jilong Xue},
  \bibinfo{person}{Youshan Miao}, \bibinfo{person}{Wei Cui},
  \bibinfo{person}{Wenxiang Hu}, \bibinfo{person}{Fan Yang},
  \bibinfo{person}{Lintao Zhang}, {and} \bibinfo{person}{Lidong Zhou}.}
  \bibinfo{year}{2020}\natexlab{}.
\newblock \bibinfo{booktitle}{\emph{RAMMER: Enabling Holistic Deep Learning
  Compiler Optimizations with Rtasks}}.
\newblock \bibinfo{publisher}{USENIX Association}, \bibinfo{address}{USA},
  \bibinfo{pages}{17}.
\newblock
\showISBNx{978-1-939133-19-9}


\bibitem[Nickolls et~al\mbox{.}(2008)]%
        {cuda}
\bibfield{author}{\bibinfo{person}{John Nickolls}, \bibinfo{person}{Ian Buck},
  \bibinfo{person}{Michael Garland}, {and} \bibinfo{person}{Kevin Skadron}.}
  \bibinfo{year}{2008}\natexlab{}.
\newblock \showarticletitle{Scalable parallel programming with cuda: Is cuda
  the parallel programming model that application developers have been waiting
  for?}
\newblock \bibinfo{journal}{\emph{Queue}} \bibinfo{volume}{6},
  \bibinfo{number}{2} (\bibinfo{year}{2008}), \bibinfo{pages}{40--53}.
\newblock


\bibitem[Nickolls and Dally(2010)]%
        {gpu}
\bibfield{author}{\bibinfo{person}{John Nickolls} {and}
  \bibinfo{person}{William~J. Dally}.} \bibinfo{year}{2010}\natexlab{}.
\newblock \showarticletitle{The GPU Computing Era}.
\newblock \bibinfo{journal}{\emph{IEEE Micro}} \bibinfo{volume}{30},
  \bibinfo{number}{2} (\bibinfo{year}{2010}), \bibinfo{pages}{56--69}.
\newblock
\urldef\tempurl%
\url{https://doi.org/10.1109/MM.2010.41}
\showDOI{\tempurl}


\bibitem[Paszke et~al\mbox{.}(2019)]%
        {pytorch}
\bibfield{author}{\bibinfo{person}{Adam Paszke}, \bibinfo{person}{Sam Gross},
  \bibinfo{person}{Francisco Massa}, \bibinfo{person}{Adam Lerer},
  \bibinfo{person}{James Bradbury}, \bibinfo{person}{Gregory Chanan},
  \bibinfo{person}{Trevor Killeen}, \bibinfo{person}{Zeming Lin},
  \bibinfo{person}{Natalia Gimelshein}, \bibinfo{person}{Luca Antiga},
  \bibinfo{person}{Alban Desmaison}, \bibinfo{person}{Andreas Kopf},
  \bibinfo{person}{Edward Yang}, \bibinfo{person}{Zachary DeVito},
  \bibinfo{person}{Martin Raison}, \bibinfo{person}{Alykhan Tejani},
  \bibinfo{person}{Sasank Chilamkurthy}, \bibinfo{person}{Benoit Steiner},
  \bibinfo{person}{Lu Fang}, \bibinfo{person}{Junjie Bai}, {and}
  \bibinfo{person}{Soumith Chintala}.} \bibinfo{year}{2019}\natexlab{}.
\newblock \showarticletitle{PyTorch: An Imperative Style, High-Performance Deep
  Learning Library}.
\newblock In \bibinfo{booktitle}{\emph{Advances in Neural Information
  Processing Systems 32}}, \bibfield{editor}{\bibinfo{person}{H.~Wallach},
  \bibinfo{person}{H.~Larochelle}, \bibinfo{person}{A.~Beygelzimer},
  \bibinfo{person}{F.~d\textquotesingle Alch\'{e}-Buc},
  \bibinfo{person}{E.~Fox}, {and} \bibinfo{person}{R.~Garnett}} (Eds.).
  \bibinfo{publisher}{Curran Associates, Inc.}, \bibinfo{pages}{8024--8035}.
\newblock
\urldef\tempurl%
\url{http://papers.neurips.cc/paper/9015-pytorch-an-imperative-style-high-performance-deep-learning-library.pdf}
\showURL{%
\tempurl}


\bibitem[Radford et~al\mbox{.}(2019)]%
        {gpt2}
\bibfield{author}{\bibinfo{person}{Alec Radford}, \bibinfo{person}{Jeffrey Wu},
  \bibinfo{person}{Rewon Child}, \bibinfo{person}{David Luan},
  \bibinfo{person}{Dario Amodei}, \bibinfo{person}{Ilya Sutskever},
  {et~al\mbox{.}}} \bibinfo{year}{2019}\natexlab{}.
\newblock \showarticletitle{Language models are unsupervised multitask
  learners}.
\newblock \bibinfo{journal}{\emph{OpenAI blog}} \bibinfo{volume}{1},
  \bibinfo{number}{8} (\bibinfo{year}{2019}), \bibinfo{pages}{9}.
\newblock


\bibitem[Ragan-Kelley et~al\mbox{.}(2013)]%
        {halide}
\bibfield{author}{\bibinfo{person}{Jonathan Ragan-Kelley},
  \bibinfo{person}{Connelly Barnes}, \bibinfo{person}{Andrew Adams},
  \bibinfo{person}{Sylvain Paris}, \bibinfo{person}{Fr{\'e}do Durand}, {and}
  \bibinfo{person}{Saman Amarasinghe}.} \bibinfo{year}{2013}\natexlab{}.
\newblock \showarticletitle{Halide: a language and compiler for optimizing
  parallelism, locality, and recomputation in image processing pipelines}. In
  \bibinfo{booktitle}{\emph{Acm Sigplan Notices}}, Vol.~\bibinfo{volume}{48}.
  ACM, \bibinfo{pages}{519--530}.
\newblock


\bibitem[Sabne(2020)]%
        {tensorflow-xla}
\bibfield{author}{\bibinfo{person}{Amit Sabne}.}
  \bibinfo{year}{2020}\natexlab{}.
\newblock \bibinfo{title}{XLA : Compiling Machine Learning for Peak
  Performance}.
\newblock
\newblock


\bibitem[Sandler et~al\mbox{.}(2018)]%
        {mobilenetv2}
\bibfield{author}{\bibinfo{person}{Mark Sandler}, \bibinfo{person}{Andrew
  Howard}, \bibinfo{person}{Menglong Zhu}, \bibinfo{person}{Andrey Zhmoginov},
  {and} \bibinfo{person}{Liang-Chieh Chen}.} \bibinfo{year}{2018}\natexlab{}.
\newblock \showarticletitle{Mobilenetv2: Inverted residuals and linear
  bottlenecks}. In \bibinfo{booktitle}{\emph{Proceedings of the IEEE conference
  on computer vision and pattern recognition}}. \bibinfo{pages}{4510--4520}.
\newblock


\bibitem[Shao et~al\mbox{.}(2022)]%
        {Shao2022TensorPO}
\bibfield{author}{\bibinfo{person}{Junru Shao}, \bibinfo{person}{Xiyou Zhou},
  \bibinfo{person}{Siyuan Feng}, \bibinfo{person}{Bohan Hou},
  \bibinfo{person}{Ruihang Lai}, \bibinfo{person}{Hongyi Jin},
  \bibinfo{person}{Wuwei Lin}, \bibinfo{person}{Masahiro Masuda},
  \bibinfo{person}{Cody~Hao Yu}, {and} \bibinfo{person}{Tianqi Chen}.}
  \bibinfo{year}{2022}\natexlab{}.
\newblock \showarticletitle{Tensor Program Optimization with Probabilistic
  Programs}.
\newblock \bibinfo{journal}{\emph{ArXiv}}  \bibinfo{volume}{abs/2205.13603}
  (\bibinfo{year}{2022}).
\newblock


\bibitem[Shen et~al\mbox{.}(2021)]%
        {nimble_haichen}
\bibfield{author}{\bibinfo{person}{Haichen Shen}, \bibinfo{person}{Jared
  Roesch}, \bibinfo{person}{Zhi Chen}, \bibinfo{person}{Wei Chen},
  \bibinfo{person}{Yong Wu}, \bibinfo{person}{Mu Li}, \bibinfo{person}{Vin
  Sharma}, \bibinfo{person}{Zachary Tatlock}, {and} \bibinfo{person}{Yida
  Wang}.} \bibinfo{year}{2021}\natexlab{}.
\newblock \showarticletitle{Nimble: Efficiently Compiling Dynamic Neural
  Networks for Model Inference}. In \bibinfo{booktitle}{\emph{Proceedings of
  Machine Learning and Systems}}, \bibfield{editor}{\bibinfo{person}{A.~Smola},
  \bibinfo{person}{A.~Dimakis}, {and} \bibinfo{person}{I.~Stoica}} (Eds.),
  Vol.~\bibinfo{volume}{3}. \bibinfo{pages}{208--222}.
\newblock
\urldef\tempurl%
\url{https://proceedings.mlsys.org/paper/2021/file/4e732ced3463d06de0ca9a15b6153677-Paper.pdf}
\showURL{%
\tempurl}


\bibitem[Sutskever et~al\mbox{.}(2014)]%
        {sutskever2014seq2seq}
\bibfield{author}{\bibinfo{person}{Ilya Sutskever}, \bibinfo{person}{Oriol
  Vinyals}, {and} \bibinfo{person}{Quoc~V Le}.}
  \bibinfo{year}{2014}\natexlab{}.
\newblock \showarticletitle{Sequence to sequence learning with neural
  networks}. In \bibinfo{booktitle}{\emph{Advances in neural information
  processing systems}}. \bibinfo{pages}{3104--3112}.
\newblock


\bibitem[Szegedy et~al\mbox{.}(2015)]%
        {szegedy2015going}
\bibfield{author}{\bibinfo{person}{Christian Szegedy}, \bibinfo{person}{Wei
  Liu}, \bibinfo{person}{Yangqing Jia}, \bibinfo{person}{Pierre Sermanet},
  \bibinfo{person}{Scott Reed}, \bibinfo{person}{Dragomir Anguelov},
  \bibinfo{person}{Dumitru Erhan}, \bibinfo{person}{Vincent Vanhoucke}, {and}
  \bibinfo{person}{Andrew Rabinovich}.} \bibinfo{year}{2015}\natexlab{}.
\newblock \showarticletitle{Going deeper with convolutions}. In
  \bibinfo{booktitle}{\emph{Proceedings of the IEEE conference on computer
  vision and pattern recognition}}. \bibinfo{pages}{1--9}.
\newblock


\bibitem[Szegedy et~al\mbox{.}(2016)]%
        {szegedy2016rethinking}
\bibfield{author}{\bibinfo{person}{Christian Szegedy}, \bibinfo{person}{Vincent
  Vanhoucke}, \bibinfo{person}{Sergey Ioffe}, \bibinfo{person}{Jon Shlens},
  {and} \bibinfo{person}{Zbigniew Wojna}.} \bibinfo{year}{2016}\natexlab{}.
\newblock \showarticletitle{Rethinking the inception architecture for computer
  vision}. In \bibinfo{booktitle}{\emph{Proceedings of the IEEE conference on
  computer vision and pattern recognition}}. \bibinfo{pages}{2818--2826}.
\newblock


\bibitem[Tang et~al\mbox{.}(2022)]%
        {freetensor}
\bibfield{author}{\bibinfo{person}{Shizhi Tang}, \bibinfo{person}{Jidong Zhai},
  \bibinfo{person}{Haojie Wang}, \bibinfo{person}{Lin Jiang},
  \bibinfo{person}{Liyan Zheng}, \bibinfo{person}{Zhenhao Yuan}, {and}
  \bibinfo{person}{Chen Zhang}.} \bibinfo{year}{2022}\natexlab{}.
\newblock \showarticletitle{FreeTensor: A Free-Form DSL with Holistic
  Optimizations for Irregular Tensor Programs}. In
  \bibinfo{booktitle}{\emph{Proceedings of the 43rd ACM SIGPLAN International
  Conference on Programming Language Design and Implementation (PLDI ’22)}}
  (San Diego, CA, USA) \emph{(\bibinfo{series}{PLDI ’22})}.
  \bibinfo{address}{New York, NY, USA}, \bibinfo{numpages}{16}~pages.
\newblock
\urldef\tempurl%
\url{https://doi.org/10.1145/3519939.3523448}
\showDOI{\tempurl}


\bibitem[Tillet et~al\mbox{.}(2019)]%
        {triton}
\bibfield{author}{\bibinfo{person}{Philippe Tillet}, \bibinfo{person}{H.~T.
  Kung}, {and} \bibinfo{person}{David Cox}.} \bibinfo{year}{2019}\natexlab{}.
\newblock \showarticletitle{Triton: An Intermediate Language and Compiler for
  Tiled Neural Network Computations}. In \bibinfo{booktitle}{\emph{Proceedings
  of the 3rd ACM SIGPLAN International Workshop on Machine Learning and
  Programming Languages}} (Phoenix, AZ, USA) \emph{(\bibinfo{series}{MAPL
  2019})}. \bibinfo{publisher}{Association for Computing Machinery},
  \bibinfo{address}{New York, NY, USA}, \bibinfo{pages}{10–19}.
\newblock
\showISBNx{9781450367196}
\urldef\tempurl%
\url{https://doi.org/10.1145/3315508.3329973}
\showDOI{\tempurl}


\bibitem[Ukarande et~al\mbox{.}(2021)]%
        {threadblock_swizzle}
\bibfield{author}{\bibinfo{person}{Aditya Ukarande}, \bibinfo{person}{Suryakant
  Patidar}, {and} \bibinfo{person}{Ram Rangan}.}
  \bibinfo{year}{2021}\natexlab{}.
\newblock \showarticletitle{Locality-Aware CTA Scheduling for Gaming
  Applications}.
\newblock \bibinfo{journal}{\emph{ACM Trans. Archit. Code Optim.}}
  \bibinfo{volume}{19}, \bibinfo{number}{1}, Article \bibinfo{articleno}{1}
  (\bibinfo{date}{dec} \bibinfo{year}{2021}), \bibinfo{numpages}{26}~pages.
\newblock
\showISSN{1544-3566}
\urldef\tempurl%
\url{https://doi.org/10.1145/3477497}
\showDOI{\tempurl}


\bibitem[Vasilache et~al\mbox{.}(2018)]%
        {tensor_comprehension}
\bibfield{author}{\bibinfo{person}{Nicolas Vasilache},
  \bibinfo{person}{Oleksandr Zinenko}, \bibinfo{person}{Theodoros Theodoridis},
  \bibinfo{person}{Priya Goyal}, \bibinfo{person}{Zach DeVito},
  \bibinfo{person}{William~S. Moses}, \bibinfo{person}{Sven Verdoolaege},
  \bibinfo{person}{Andrew Adams}, {and} \bibinfo{person}{Albert Cohen}.}
  \bibinfo{year}{2018}\natexlab{}.
\newblock \showarticletitle{Tensor Comprehensions: Framework-Agnostic
  High-Performance Machine Learning Abstractions}.
\newblock \bibinfo{journal}{\emph{ArXiv}}  \bibinfo{volume}{abs/1802.04730}
  (\bibinfo{year}{2018}).
\newblock


\bibitem[Vaswani et~al\mbox{.}(2017)]%
        {attention}
\bibfield{author}{\bibinfo{person}{Ashish Vaswani}, \bibinfo{person}{Noam
  Shazeer}, \bibinfo{person}{Niki Parmar}, \bibinfo{person}{Jakob Uszkoreit},
  \bibinfo{person}{Llion Jones}, \bibinfo{person}{Aidan~N Gomez},
  \bibinfo{person}{\L~ukasz Kaiser}, {and} \bibinfo{person}{Illia Polosukhin}.}
  \bibinfo{year}{2017}\natexlab{}.
\newblock \showarticletitle{Attention is All you Need}. In
  \bibinfo{booktitle}{\emph{Advances in Neural Information Processing
  Systems}}, \bibfield{editor}{\bibinfo{person}{I.~Guyon},
  \bibinfo{person}{U.~Von Luxburg}, \bibinfo{person}{S.~Bengio},
  \bibinfo{person}{H.~Wallach}, \bibinfo{person}{R.~Fergus},
  \bibinfo{person}{S.~Vishwanathan}, {and} \bibinfo{person}{R.~Garnett}}
  (Eds.), Vol.~\bibinfo{volume}{30}. \bibinfo{publisher}{Curran Associates,
  Inc.}
\newblock
\urldef\tempurl%
\url{https://proceedings.neurips.cc/paper/2017/file/3f5ee243547dee91fbd053c1c4a845aa-Paper.pdf}
\showURL{%
\tempurl}


\bibitem[Wang et~al\mbox{.}(2021)]%
        {pet}
\bibfield{author}{\bibinfo{person}{Haojie Wang}, \bibinfo{person}{Jidong Zhai},
  \bibinfo{person}{Mingyu Gao}, \bibinfo{person}{Zixuan Ma},
  \bibinfo{person}{Shizhi Tang}, \bibinfo{person}{Liyan Zheng},
  \bibinfo{person}{Yuanzhi Li}, \bibinfo{person}{Kaiyuan Rong},
  \bibinfo{person}{Yuanyong Chen}, {and} \bibinfo{person}{Zhihao Jia}.}
  \bibinfo{year}{2021}\natexlab{}.
\newblock \showarticletitle{PET: Optimizing Tensor Programs with Partially
  Equivalent Transformations and Automated Corrections}. In
  \bibinfo{booktitle}{\emph{USENIX Symposium on Operating Systems Design and
  Implementation}}.
\newblock


\bibitem[Weng et~al\mbox{.}(2021)]%
        {unit}
\bibfield{author}{\bibinfo{person}{Jian Weng}, \bibinfo{person}{Animesh Jain},
  \bibinfo{person}{Jie Wang}, \bibinfo{person}{Leyuan Wang},
  \bibinfo{person}{Yida Wang}, {and} \bibinfo{person}{Tony Nowatzki}.}
  \bibinfo{year}{2021}\natexlab{}.
\newblock \bibinfo{booktitle}{\emph{UNIT: Unifying Tensorized Instruction
  Compilation}}.
\newblock \bibinfo{publisher}{IEEE Press}, \bibinfo{pages}{77–89}.
\newblock
\showISBNx{9781728186139}
\urldef\tempurl%
\url{https://doi.org/10.1109/CGO51591.2021.9370330}
\showURL{%
\tempurl}


\bibitem[Xing et~al\mbox{.}(2022)]%
        {bolt}
\bibfield{author}{\bibinfo{person}{Jiarong Xing}, \bibinfo{person}{Leyuan
  Wang}, \bibinfo{person}{Shang Zhang}, \bibinfo{person}{Jack Chen},
  \bibinfo{person}{Ang Chen}, {and} \bibinfo{person}{Yibo Zhu}.}
  \bibinfo{year}{2022}\natexlab{}.
\newblock \showarticletitle{Bolt: Bridging the Gap between Auto-tuners and
  Hardware-native Performance}. In \bibinfo{booktitle}{\emph{Proceedings of
  Machine Learning and Systems}}, Vol.~\bibinfo{volume}{4}.
\newblock


\bibitem[Xu et~al\mbox{.}(2022)]%
        {aitemplate}
\bibfield{author}{\bibinfo{person}{Bing Xu}, \bibinfo{person}{Ying Zhang},
  \bibinfo{person}{Hao Lu}, \bibinfo{person}{Yang Chen}, \bibinfo{person}{Terry
  Chen}, \bibinfo{person}{Mike Iovine}, \bibinfo{person}{Mu-Chu Lee}, {and}
  \bibinfo{person}{Zhijing Li}.} \bibinfo{year}{2022}\natexlab{}.
\newblock \bibinfo{booktitle}{\emph{{AITemplate}}}.
\newblock
\urldef\tempurl%
\url{https://github.com/facebookincubator/AITemplate}
\showURL{%
\tempurl}


\bibitem[Yang et~al\mbox{.}(2021)]%
        {tensat}
\bibfield{author}{\bibinfo{person}{Yichen Yang}, \bibinfo{person}{Phitchaya
  Phothilimthana}, \bibinfo{person}{Yisu Wang}, \bibinfo{person}{Max Willsey},
  \bibinfo{person}{Sudip Roy}, {and} \bibinfo{person}{Jacques Pienaar}.}
  \bibinfo{year}{2021}\natexlab{}.
\newblock \showarticletitle{Equality Saturation for Tensor Graph
  Superoptimization}. In \bibinfo{booktitle}{\emph{Proceedings of Machine
  Learning and Systems}}, \bibfield{editor}{\bibinfo{person}{A.~Smola},
  \bibinfo{person}{A.~Dimakis}, {and} \bibinfo{person}{I.~Stoica}} (Eds.),
  Vol.~\bibinfo{volume}{3}. \bibinfo{pages}{255--268}.
\newblock
\urldef\tempurl%
\url{https://proceedings.mlsys.org/paper/2021/file/65ded5353c5ee48d0b7d48c591b8f430-Paper.pdf}
\showURL{%
\tempurl}


\bibitem[Zhao et~al\mbox{.}(2022)]%
        {apollo}
\bibfield{author}{\bibinfo{person}{Jie Zhao}, \bibinfo{person}{Xiong Gao},
  \bibinfo{person}{Ruijie Xia}, \bibinfo{person}{Zhaochuang Zhang},
  \bibinfo{person}{Deshi Chen}, \bibinfo{person}{Lei Chen},
  \bibinfo{person}{Renwei Zhang}, \bibinfo{person}{Zhen Geng},
  \bibinfo{person}{Bin Cheng}, {and} \bibinfo{person}{Xuefeng Jin}.}
  \bibinfo{year}{2022}\natexlab{}.
\newblock \showarticletitle{Apollo: Automatic Partition-based Operator Fusion
  through Layer by Layer Optimization}. In
  \bibinfo{booktitle}{\emph{Proceedings of Machine Learning and Systems}},
  \bibfield{editor}{\bibinfo{person}{D.~Marculescu}, \bibinfo{person}{Y.~Chi},
  {and} \bibinfo{person}{C.~Wu}} (Eds.), Vol.~\bibinfo{volume}{4}.
  \bibinfo{pages}{1--19}.
\newblock
\urldef\tempurl%
\url{https://proceedings.mlsys.org/paper/2022/file/069059b7ef840f0c74a814ec9237b6ec-Paper.pdf}
\showURL{%
\tempurl}


\bibitem[Zhao et~al\mbox{.}(2021)]%
        {akg}
\bibfield{author}{\bibinfo{person}{Jie Zhao}, \bibinfo{person}{Bojie Li},
  \bibinfo{person}{Wang Nie}, \bibinfo{person}{Zhenglin Geng},
  \bibinfo{person}{Renwei Zhang}, \bibinfo{person}{Xiong Gao},
  \bibinfo{person}{Bin Cheng}, \bibinfo{person}{Chen Wu}, \bibinfo{person}{Yun
  Cheng}, \bibinfo{person}{Zheng Li}, \bibinfo{person}{Peng Di},
  \bibinfo{person}{Kun Zhang}, {and} \bibinfo{person}{Xuefeng Jin}.}
  \bibinfo{year}{2021}\natexlab{}.
\newblock \showarticletitle{AKG: automatic kernel generation for neural
  processing units using polyhedral transformations}.
\newblock \bibinfo{journal}{\emph{Proceedings of the 42nd ACM SIGPLAN
  International Conference on Programming Language Design and Implementation}}
  (\bibinfo{year}{2021}).
\newblock


\bibitem[Zheng et~al\mbox{.}(2022b)]%
        {dietcode}
\bibfield{author}{\bibinfo{person}{Bojian Zheng}, \bibinfo{person}{Ziheng
  Jiang}, \bibinfo{person}{Cody~Hao Yu}, \bibinfo{person}{Haichen Shen},
  \bibinfo{person}{Joshua Fromm}, \bibinfo{person}{Yizhi Liu},
  \bibinfo{person}{Yida Wang}, \bibinfo{person}{Luis Ceze},
  \bibinfo{person}{Tianqi Chen}, {and} \bibinfo{person}{Gennady Pekhimenko}.}
  \bibinfo{year}{2022}\natexlab{b}.
\newblock \showarticletitle{{DietCode}: Automatic Optimization for Dynamic
  Tensor Programs}. In \bibinfo{booktitle}{\emph{Proceedings of Machine
  Learning and Systems}}, \bibfield{editor}{\bibinfo{person}{D.~Marculescu},
  \bibinfo{person}{Y.~Chi}, {and} \bibinfo{person}{C.~Wu}} (Eds.),
  Vol.~\bibinfo{volume}{4}. \bibinfo{pages}{848--863}.
\newblock
\urldef\tempurl%
\url{https://proceedings.mlsys.org/paper/2022/file/fa7cdfad1a5aaf8370ebeda47a1ff1c3-Paper.pdf}
\showURL{%
\tempurl}


\bibitem[Zheng et~al\mbox{.}(2020c)]%
        {echo}
\bibfield{author}{\bibinfo{person}{Bojian Zheng}, \bibinfo{person}{Nandita
  Vijaykumar}, {and} \bibinfo{person}{Gennady Pekhimenko}.}
  \bibinfo{year}{2020}\natexlab{c}.
\newblock \showarticletitle{Echo: Compiler-Based GPU Memory Footprint Reduction
  for LSTM RNN Training}. In \bibinfo{booktitle}{\emph{Proceedings of the
  ACM/IEEE 47th Annual International Symposium on Computer Architecture}}
  (Virtual Event) \emph{(\bibinfo{series}{ISCA '20})}. \bibinfo{publisher}{IEEE
  Press}, \bibinfo{pages}{1089–1102}.
\newblock
\showISBNx{9781728146614}
\urldef\tempurl%
\url{https://doi.org/10.1109/ISCA45697.2020.00092}
\showDOI{\tempurl}


\bibitem[Zheng et~al\mbox{.}(2020a)]%
        {ansor}
\bibfield{author}{\bibinfo{person}{Lianmin Zheng}, \bibinfo{person}{Chengfan
  Jia}, \bibinfo{person}{Minmin Sun}, \bibinfo{person}{Zhao Wu},
  \bibinfo{person}{Cody~Hao Yu}, \bibinfo{person}{Ameer Haj-Ali},
  \bibinfo{person}{Yida Wang}, \bibinfo{person}{Jun Yang},
  \bibinfo{person}{Danyang Zhuo}, \bibinfo{person}{Koushik Sen},
  \bibinfo{person}{Joseph~E. Gonzalez}, {and} \bibinfo{person}{Ion Stoica}.}
  \bibinfo{year}{2020}\natexlab{a}.
\newblock \showarticletitle{Ansor: Generating High-Performance Tensor Programs
  for Deep Learning}. In \bibinfo{booktitle}{\emph{14th USENIX Symposium on
  Operating Systems Design and Implementation (OSDI 20)}}.
  \bibinfo{pages}{863--879}.
\newblock


\bibitem[Zheng et~al\mbox{.}(2022a)]%
        {AMOS}
\bibfield{author}{\bibinfo{person}{Size Zheng}, \bibinfo{person}{Renze Chen},
  \bibinfo{person}{Anjiang Wei}, \bibinfo{person}{Yicheng Jin},
  \bibinfo{person}{Qin Han}, \bibinfo{person}{Liqiang Lu},
  \bibinfo{person}{Bingyang Wu}, \bibinfo{person}{Xiuhong Li},
  \bibinfo{person}{Shengen Yan}, {and} \bibinfo{person}{Yun Liang}.}
  \bibinfo{year}{2022}\natexlab{a}.
\newblock \showarticletitle{AMOS: Enabling automatic mapping for Tensor
  Computations on spatial Accelerators with Hardware Abstraction}. In
  \bibinfo{booktitle}{\emph{Proceedings of the 49th Annual International
  Symposium on Computer Architecture}} (New York, New York)
  \emph{(\bibinfo{series}{ISCA '22})}. \bibinfo{publisher}{Association for
  Computing Machinery}, \bibinfo{address}{New York, NY, USA},
  \bibinfo{pages}{874–887}.
\newblock
\showISBNx{9781450386104}
\urldef\tempurl%
\url{https://doi.org/10.1145/3470496.3527440}
\showDOI{\tempurl}


\bibitem[Zheng et~al\mbox{.}(2020b)]%
        {flextensor}
\bibfield{author}{\bibinfo{person}{Size Zheng}, \bibinfo{person}{Yun Liang},
  \bibinfo{person}{Shuo Wang}, \bibinfo{person}{Renze Chen}, {and}
  \bibinfo{person}{Kaiwen Sheng}.} \bibinfo{year}{2020}\natexlab{b}.
\newblock \showarticletitle{FlexTensor: An Automatic Schedule Exploration and
  Optimization Framework for Tensor Computation on Heterogeneous System}.
\newblock \bibinfo{journal}{\emph{Proceedings of the Twenty-Fifth International
  Conference on Architectural Support for Programming Languages and Operating
  Systems}} (\bibinfo{year}{2020}).
\newblock


\bibitem[Zheng et~al\mbox{.}(2022c)]%
        {astitch}
\bibfield{author}{\bibinfo{person}{Zhen Zheng}, \bibinfo{person}{Xuanda Yang},
  \bibinfo{person}{Pengzhan Zhao}, \bibinfo{person}{Guoping Long},
  \bibinfo{person}{Kai Zhu}, \bibinfo{person}{Feiwen Zhu},
  \bibinfo{person}{Wenyi Zhao}, \bibinfo{person}{Xiaoyong Liu},
  \bibinfo{person}{Jun Yang}, \bibinfo{person}{Jidong Zhai},
  \bibinfo{person}{Shuaiwen~Leon Song}, {and} \bibinfo{person}{Wei Lin}.}
  \bibinfo{year}{2022}\natexlab{c}.
\newblock \showarticletitle{AStitch: Enabling a New Multi-Dimensional
  Optimization Space for Memory-Intensive ML Training and Inference on Modern
  SIMT Architectures}. In \bibinfo{booktitle}{\emph{Proceedings of the 27th ACM
  International Conference on Architectural Support for Programming Languages
  and Operating Systems}} (Lausanne, Switzerland)
  \emph{(\bibinfo{series}{ASPLOS '22})}. \bibinfo{publisher}{Association for
  Computing Machinery}, \bibinfo{address}{New York, NY, USA},
  \bibinfo{pages}{359–373}.
\newblock
\showISBNx{9781450392051}
\urldef\tempurl%
\url{https://doi.org/10.1145/3503222.3507723}
\showDOI{\tempurl}


\bibitem[Zhu et~al\mbox{.}(2022)]%
        {roller}
\bibfield{author}{\bibinfo{person}{Hongyu Zhu}, \bibinfo{person}{Ruofan Wu},
  \bibinfo{person}{Yijia Diao}, \bibinfo{person}{Shanbin Ke},
  \bibinfo{person}{Haoyu Li}, \bibinfo{person}{Chen Zhang},
  \bibinfo{person}{Jilong Xue}, \bibinfo{person}{Lingxiao Ma},
  \bibinfo{person}{Yuqing Xia}, \bibinfo{person}{Wei Cui}, \bibinfo{person}{Fan
  Yang}, \bibinfo{person}{Mao Yang}, \bibinfo{person}{Lidong Zhou},
  \bibinfo{person}{Asaf Cidon}, {and} \bibinfo{person}{Gennady Pekhimenko}.}
  \bibinfo{year}{2022}\natexlab{}.
\newblock \showarticletitle{{ROLLER}: Fast and Efficient Tensor Compilation for
  Deep Learning}. In \bibinfo{booktitle}{\emph{16th USENIX Symposium on
  Operating Systems Design and Implementation (OSDI 22)}}.
  \bibinfo{publisher}{USENIX Association}, \bibinfo{address}{Carlsbad, CA},
  \bibinfo{pages}{233--248}.
\newblock
\showISBNx{978-1-939133-28-1}
\urldef\tempurl%
\url{https://www.usenix.org/conference/osdi22/presentation/zhu}
\showURL{%
\tempurl}


\bibitem[Zhu et~al\mbox{.}(2021)]%
        {disc}
\bibfield{author}{\bibinfo{person}{K. Zhu}, \bibinfo{person}{W.Y. Zhao},
  \bibinfo{person}{Z. Zheng}, \bibinfo{person}{T.Y. Guo}, \bibinfo{person}{P.Z.
  Zhao}, \bibinfo{person}{J.J. Bai}, \bibinfo{person}{J. Yang},
  \bibinfo{person}{X.Y. Liu}, \bibinfo{person}{L.S. Diao}, {and}
  \bibinfo{person}{W. Lin}.} \bibinfo{year}{2021}\natexlab{}.
\newblock \showarticletitle{DISC: A Dynamic Shape Compiler for Machine Learning
  Workloads}. In \bibinfo{booktitle}{\emph{Proceedings of the 1st Workshop on
  Machine Learning and Systems}} (Online, United Kingdom)
  \emph{(\bibinfo{series}{EuroMLSys '21})}. \bibinfo{publisher}{Association for
  Computing Machinery}, \bibinfo{address}{New York, NY, USA},
  \bibinfo{pages}{89–95}.
\newblock
\showISBNx{9781450382984}
\urldef\tempurl%
\url{https://doi.org/10.1145/3437984.3458838}
\showDOI{\tempurl}


\bibitem[Zoph and Le(2016)]%
        {rl_nas}
\bibfield{author}{\bibinfo{person}{Barret Zoph} {and} \bibinfo{person}{Quoc~V
  Le}.} \bibinfo{year}{2016}\natexlab{}.
\newblock \showarticletitle{Neural architecture search with reinforcement
  learning}.
\newblock \bibinfo{journal}{\emph{arXiv preprint arXiv:1611.01578}}
  (\bibinfo{year}{2016}).
\newblock


\end{thebibliography}

\end{document}